\documentclass[acmtog]{acmart}

\usepackage{booktabs}
\usepackage{caption}
\usepackage{multirow}

\usepackage{epsfig}
\usepackage{graphicx}
\usepackage{makecell}
\usepackage{wrapfig}

\usepackage{pifont}
\usepackage{booktabs} 

\usepackage[ruled]{algorithm2e} 

\SetAlFnt{\small}
\SetAlCapFnt{\small}
\SetAlCapNameFnt{\small}
\SetAlCapHSkip{0pt}

\acmJournal{TOG}

\usepackage{overpic}
\usepackage{enumitem} 
\usepackage{overpic} 
\usepackage{color}

\definecolor{turquoise}{cmyk}{0.65,0,0.1,0.3}
\definecolor{purple}{rgb}{0.65,0,0.65}
\definecolor{dark_green}{rgb}{0, 0.5, 0}
\definecolor{orange}{rgb}{0.8, 0.6, 0.2}
\definecolor{red}{rgb}{0.8, 0.2, 0.2}
\definecolor{darkred}{rgb}{0.6, 0.1, 0.05}
\definecolor{blueish}{rgb}{0.0, 0.3, .6}
\definecolor{light_gray}{rgb}{0.7, 0.7, .7}
\definecolor{pink}{rgb}{1, 0, 1}
\definecolor{greyblue}{rgb}{0.25, 0.25, 1}

\newcommand\cnum[1]{\raisebox{.5pt}{\textcircled{\raisebox{-0.9pt}{#1}}}}

\usepackage{blindtext}

\begin{document}
\title{FashionEngine: Interactive 3D Human Generation and Editing via Multimodal Controls}

\author{Tao Hu}
\affiliation{%
  \institution{S-Lab, Nanyang Technological University}
  \country{Singapore}
}

\author{Fangzhou Hong}
\affiliation{%
  \institution{S-Lab, Nanyang Technological University}
  \country{Singapore}
}

\author{Zhaoxi Chen}
\affiliation{%
  \institution{S-Lab, Nanyang Technological University}
  \country{Singapore}
}

\author{Ziwei Liu}
\affiliation{%
  \institution{S-Lab, Nanyang Technological University}
  \country{Singapore}
}

\newcommand{\nickname}{FashionEngine}
\newcommand{\etal}{\emph{et al.}\xspace}
\newcommand{\ie}{\emph{i.e.}\xspace}
\newcommand{\eg}{\emph{e.g.}\xspace}

\newcommand{\tabSummary}{
\begin{table}[t] \small    
    \renewcommand{\arraystretch}{1.2}
    \centering
    \caption{A set of recent generation and editing approaches.}
    \setlength{\tabcolsep}{0.9mm}{
    \begin{tabular}{lccccc}
       \hline
        Methods             & Uncond.  & Text   & Image  & Sketch  & 3D-aware \\ \hline
        EG3D~\scriptsize{\cite{eg3d}}    & \checkmark   &   &   &    & \checkmark     \\ 
        StyleSDF~\scriptsize{\cite{stylesdf}} & \checkmark   &   &   &    & \checkmark     \\ 
        EVA3D~\scriptsize{\cite{eva}}    & \checkmark   &   &   &    & \checkmark     \\  
        AG3D~\scriptsize{\cite{ag3d}}     & \checkmark   &   &   &    & \checkmark     \\
        StructLDM~\scriptsize{\cite{structldm}}     & \checkmark   &   &   &    & \checkmark     \\
        \hline
        DragGAN~\scriptsize{\cite{pan2023drag}} & \checkmark   &   &   &  \checkmark  &      \\
        InstructP2P~\scriptsize{\cite{brooks2022instructpix2pix}} & \checkmark   & \checkmark   &   &   &      \\

        Text2Human~\scriptsize{\cite{jiang2022text2human}} &   & \checkmark   &   &   &      \\
        Text2Performer~\scriptsize{\cite{jiang2023text2performer}} &   & \checkmark   &   &   &      \\        
        \hline
        FashionEngine (Ours) & \checkmark  & \checkmark   & \checkmark  & \checkmark  &  \checkmark    \\ \hline
    \end{tabular}}
    \label{tab:summary}
    \end{table}
}

\newcommand{\tabAbMixer}{
\begin{table}[t] 
    \small
    \renewcommand{\arraystretch}{1.3}
    \centering
    \caption{Ablation study of the size of Receptive Field (RF) in global mixer.} 
   \vspace{-0.04in}
    \begin{tabular}{lcccc}
        \hline
         & LPIPS $\downarrow$ & FID $\downarrow$ & PSNR $\uparrow$ \\ \hline
        $RF = 2$ & .067 & 14.880 & 23.575 \\ \hline
        $RF = 4$ & .060 & 12.077 & 23.803 \\ \hline
    \end{tabular}       
    \label{tab:AbMixer}
    \end{table}
}

\newcommand{\tabclipscore}{
\begin{table}[t]
    \small    
    \renewcommand{\arraystretch}{1.2}
    \centering
    \caption{Quantitative Evaluation for Text-Driven Editing.}
    \vspace{-0.04in}
    \setlength{\tabcolsep}{0.9mm}{
    \begin{tabular}{lcc}
        \hline
        Metrics $\uparrow$ & InstructPix2Pix & Ours \\
        CLIP Score (ViT-B/32) $\times 100$ & 25.42 & \textbf{26.32} \\
        CLIP Score (ViT-B/16) $\times 100$ & 25.63 & \textbf{26.34} \\
        CLIP Score (ViT-L/16) $\times 100$ & 18.54 & \textbf{20.25} \\
        \hline
        Identity Preservation $\times 100$ & 25.36 & \textbf{89.83} \\
        \hline
        Appearance Preservation (PSNR) & 28.55 & \textbf{33.90} \\
        \hline
    \end{tabular}}
    \label{tab:clipscore}
    \end{table}
}

\newcommand{\figMPrior}{
\begin{figure}
    \includegraphics[width=.48\textwidth]{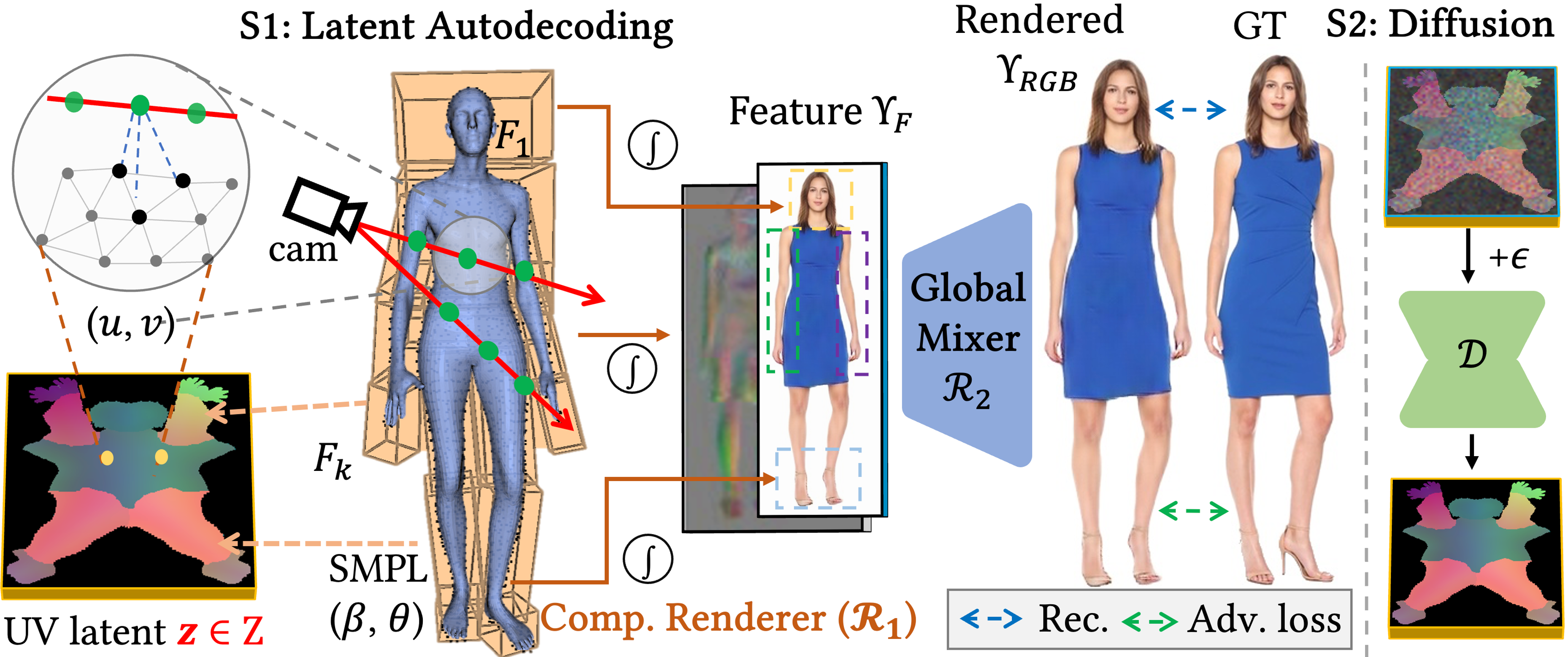}
		\vspace{-0.08in}          
    \caption{3D human prior learning \cite{structldm} in two stages (S1 and S2). S1 learns an auto-decoder containing a set of structured embeddings ${Z}$ corresponding to the human subjects in the training dataset. The embeddings ${Z}$ are then employed to train a latent diffusion model in the semantic UV latent space in the second stage.}
		\vspace{-0.08in}  
    \label{fig:m1prior}
\end{figure}
}

\newcommand{\figMSpace}{
\begin{figure}
    \includegraphics[width=.48\textwidth]{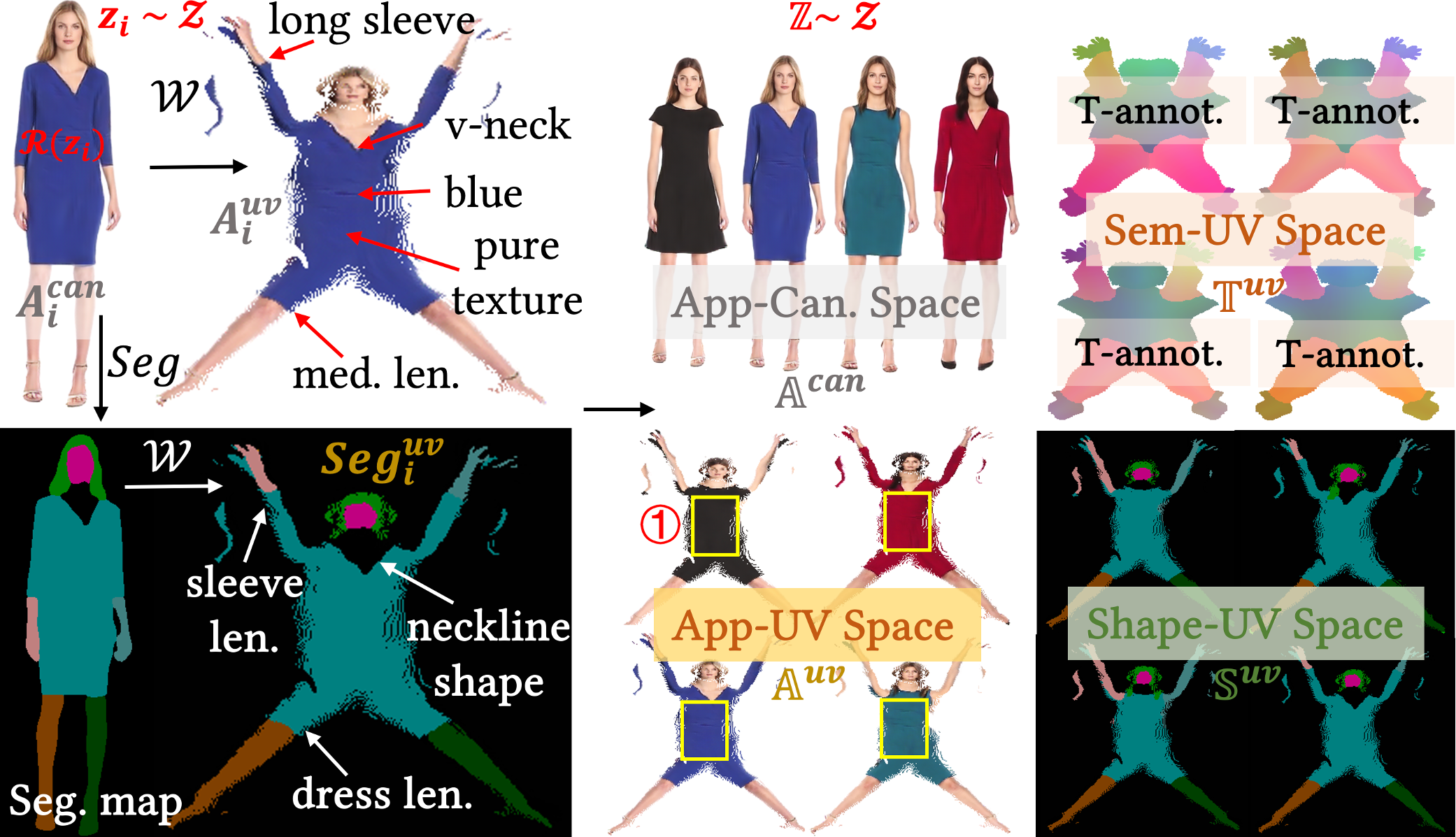}   
		\vspace{-0.08in}          
    \caption{Multimodality-UV Space (Sec. \ref{sec:method_space}). Based on the learned prior $\mathcal{Z}$, we construct a Multimodality-UV space including an Appearance-Canonical Space (App-Can, $\mathbb{A}^{can}$), an Appearance-UV Space (App-UV, $\mathbb{A}^{uv}$), a textual Semantics-UV Space (Sem-UV, $\mathbb{T}^{uv}$), and Shape-UV Space ($\mathbb{S}^{uv}$).
		}
		\vspace{-0.08in}  
    \label{fig:m2space}
\end{figure}
}

\newcommand{\figMGeneration}{
\begin{figure*}
    \includegraphics[width=\textwidth]{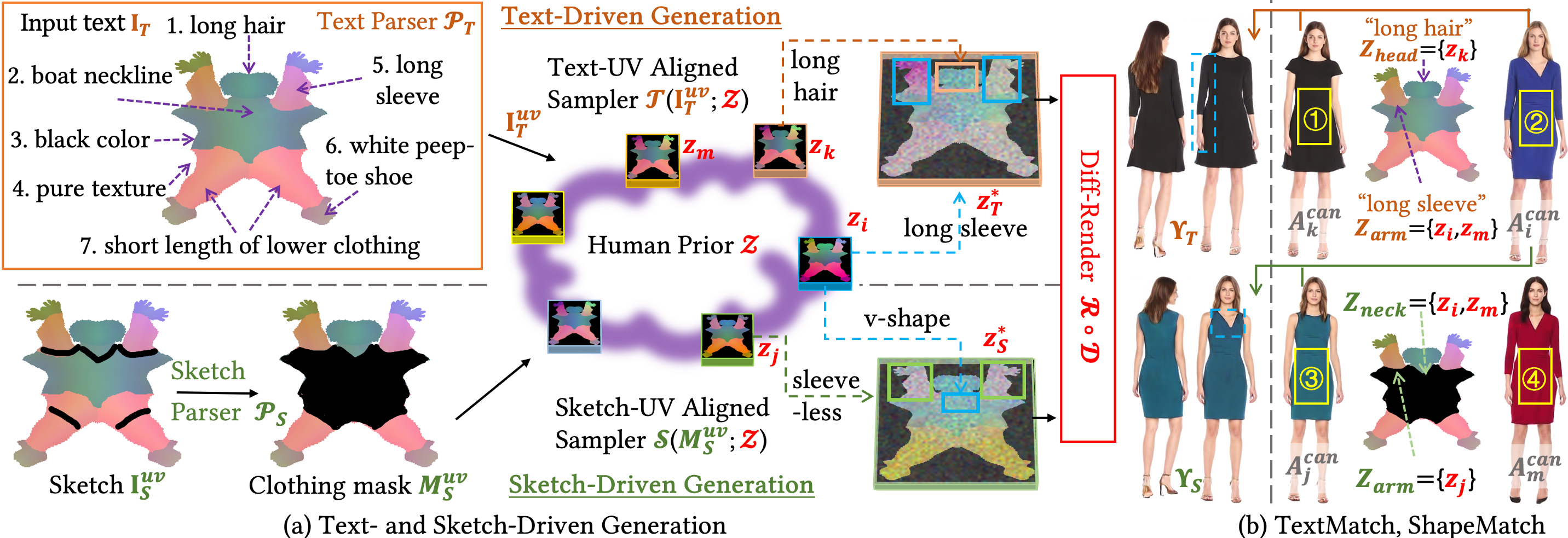}
		\vspace{-0.08in}             
    \caption{Pipeline of multimodal generation Sec. \ref{sec:method_generation}.  (a) Text- and sketch-driven generation: Given text input $\mathbf{I}_T$ or sketch input $\mathbf{I}^{uv}_S$ in the template UV space, we present Text-UV Aligned Samplers and Sketch-UV Aligned Samplers to sample latents ($z^{*}_{T}$ and $z^{*}_{S}$) from the learned human prior $\mathcal{Z}$ (Sec \ref{sec:method_prior}) respectively, which can be rendered into images by latent diffusion and rendering (Diff-Render) \cite{structldm}. (b) Illustration of TextMatch and ShapeMatch: $\{z_k, z_i\}$ $\cnum{1}\cnum{2}$ and $\{z_j, z_i\}$ $\cnum{3}\cnum{2}$ are taken as the best match to construct the target latents ($z^{*}_{T}$ and $z^{*}_{S}$) for the text or sketch input based on the TextMatch and ShapeMatch algorithms respectively.}
    \label{fig:m2gen}
\end{figure*}
}

\newcommand{\figMEdit}{
\begin{figure}
    \includegraphics[width=.48\textwidth]{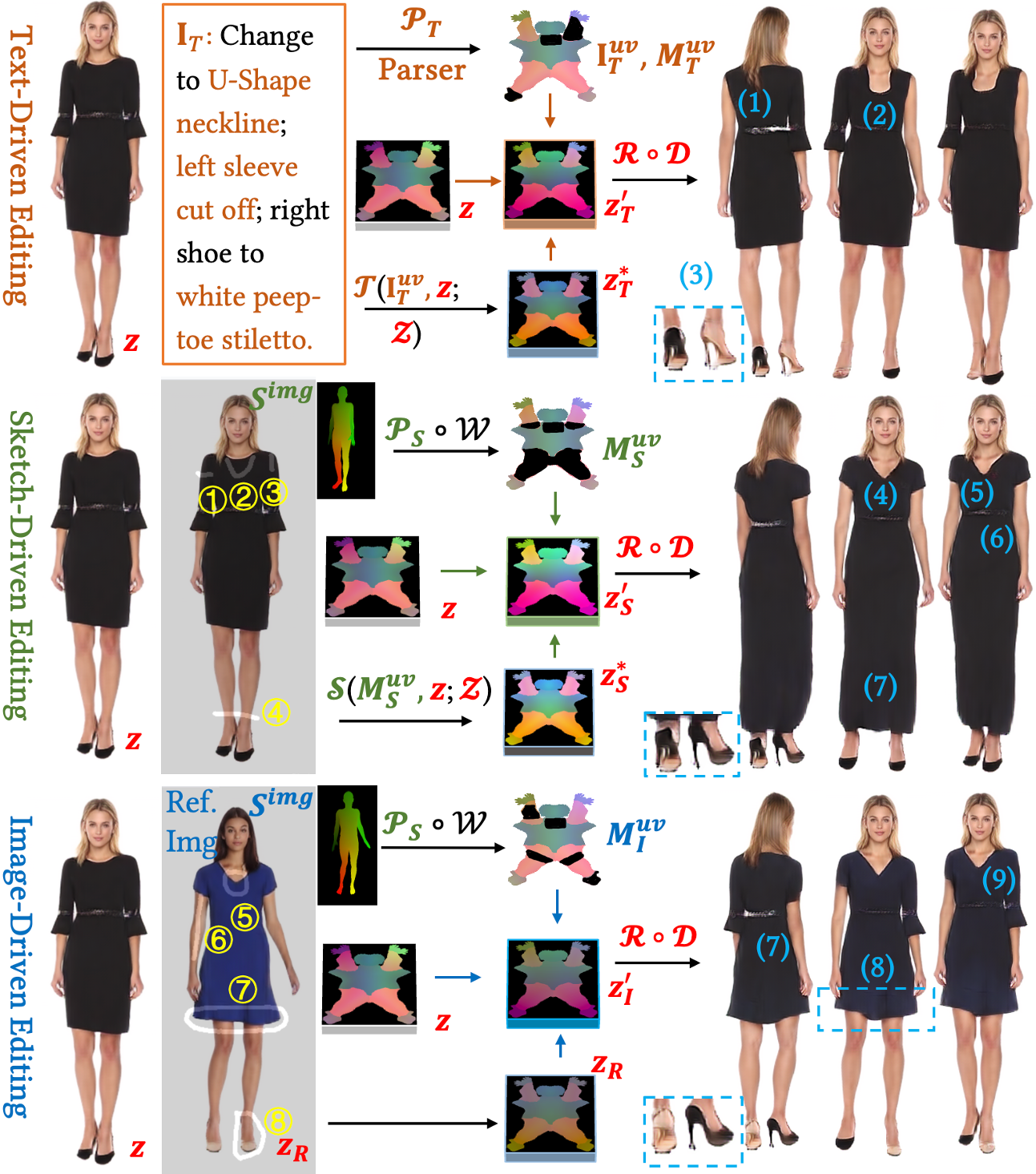}
		\vspace{-0.08in}                    
    \caption{Text-, Sketch-, and Image-Driven Editing (Sec. \ref{sec:method_editing}). To edit a source human with latent $z$, \nickname{} allows users to type texts $\mathbf{I}_T$,  draw sketches $S^{img}$, or provide a reference image with sketh masks for style transfer, and the target latents are constructed corresponding to the user inputs. Note that the sketch can describe the length of sleeves in two different ways (\eg, \cnum{1}\cnum{3}), or  describe the geometry (\eg, \cnum{2}).}
    \label{fig:m3edit}
\end{figure}
}

\newcommand{\figAb}{
\begin{figure}
    \includegraphics[width=.48\textwidth]{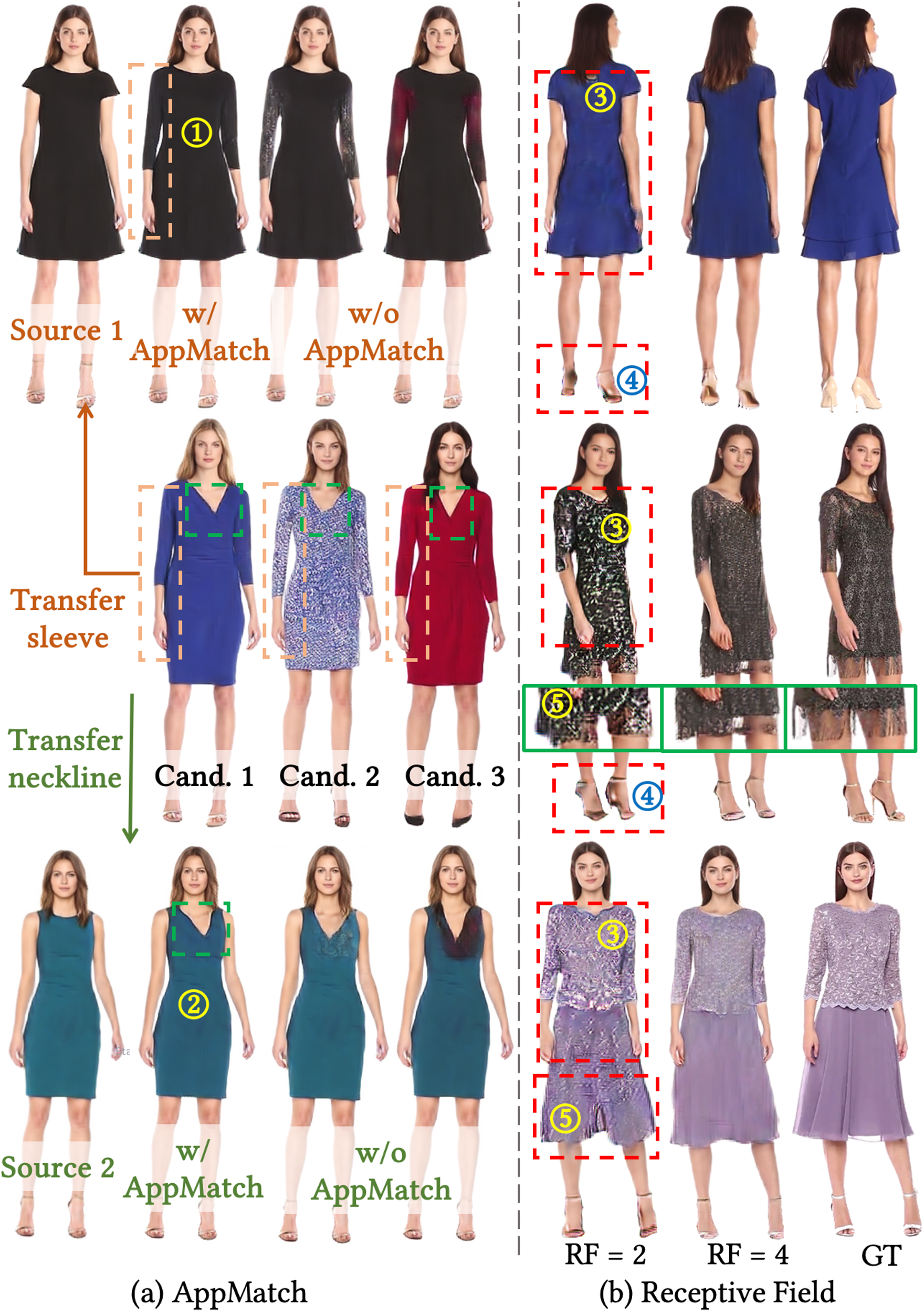}          	\vspace{-0.24in}
    \caption{Ablation study of {AppMatch} (a) and the size of Receptive Field (b).}
    \label{fig:ab}
\end{figure}
}
  
\newcommand{\figUserStudy}{
\begin{figure}[]
	\begin{center}
        \includegraphics[width=\linewidth]{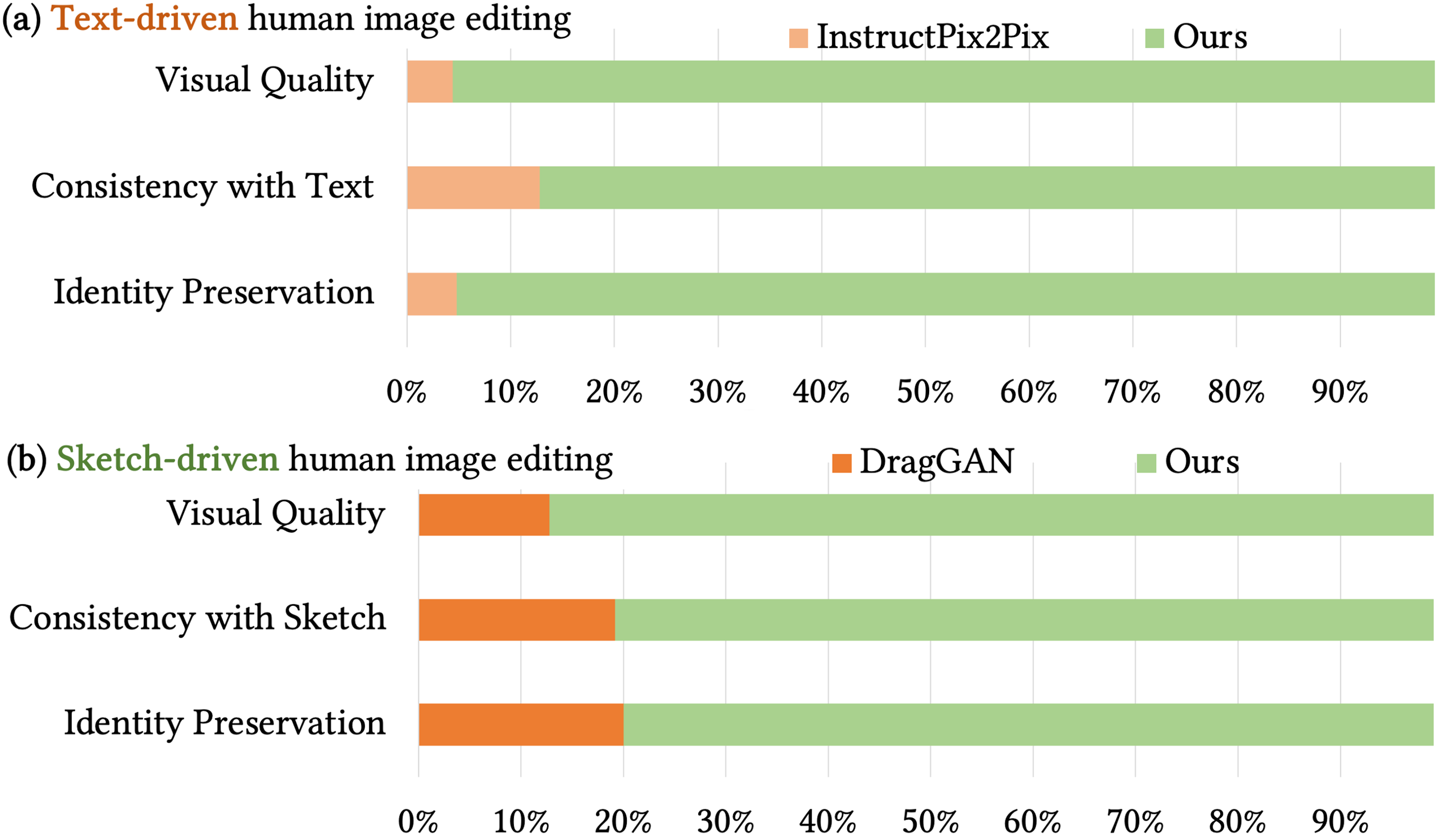}
	\end{center}
	\vspace{-0.08in}
	\caption{User study on conditional human image editing quantitatively shows the superiority of \nickname{} over text-driven baseline InstructPix2Pix \cite{brooks2022instructpix2pix} (a), and sketch-driven baseline DragGAN \cite{pan2023drag} (b) in three aspects: 1) visual quality, 2) consistency with input, and 3) identity reservation. }
	\label{fig:userstudy}
\end{figure}
}

\newcommand{\figedit}{
\begin{figure*}[]
	\begin{center}
		\includegraphics[width=\textwidth]{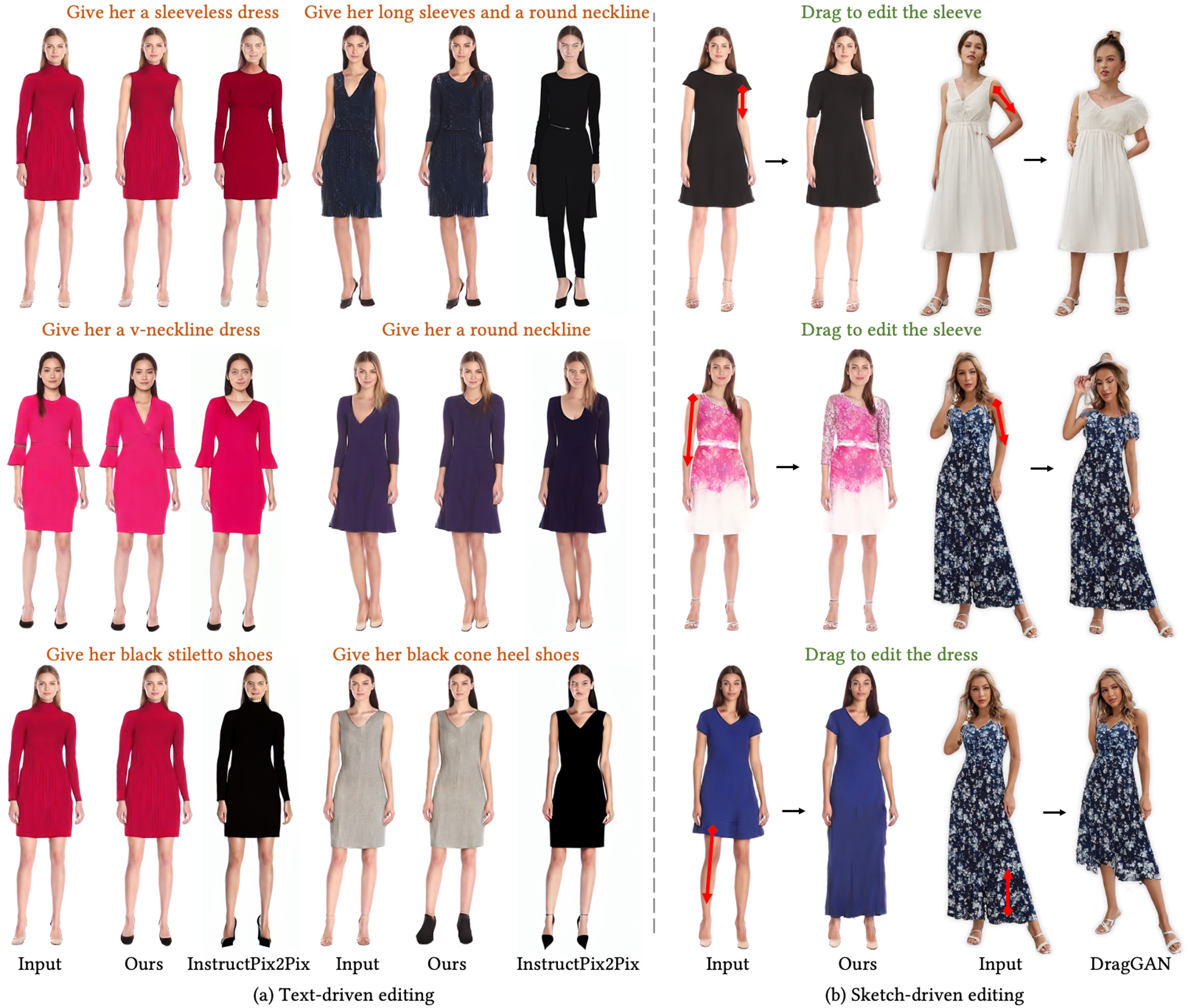}   
          \end{center}
    \vspace{-0.08in}
  
	\caption{Qualitative comparisons with InstructPix2Pix \cite{brooks2022instructpix2pix} and DragGAN \cite{pan2023drag} for text-driven and sketch-driven editing. We generate high-quality images with faithful identity preservation, and the editing results are well-aligned with the text/sketch inputs.} 
	\label{fig:edit}

\end{figure*}
}

\newcommand{\figUI}{
\begin{figure*}[]
	\begin{center}
		\includegraphics[width=\textwidth]{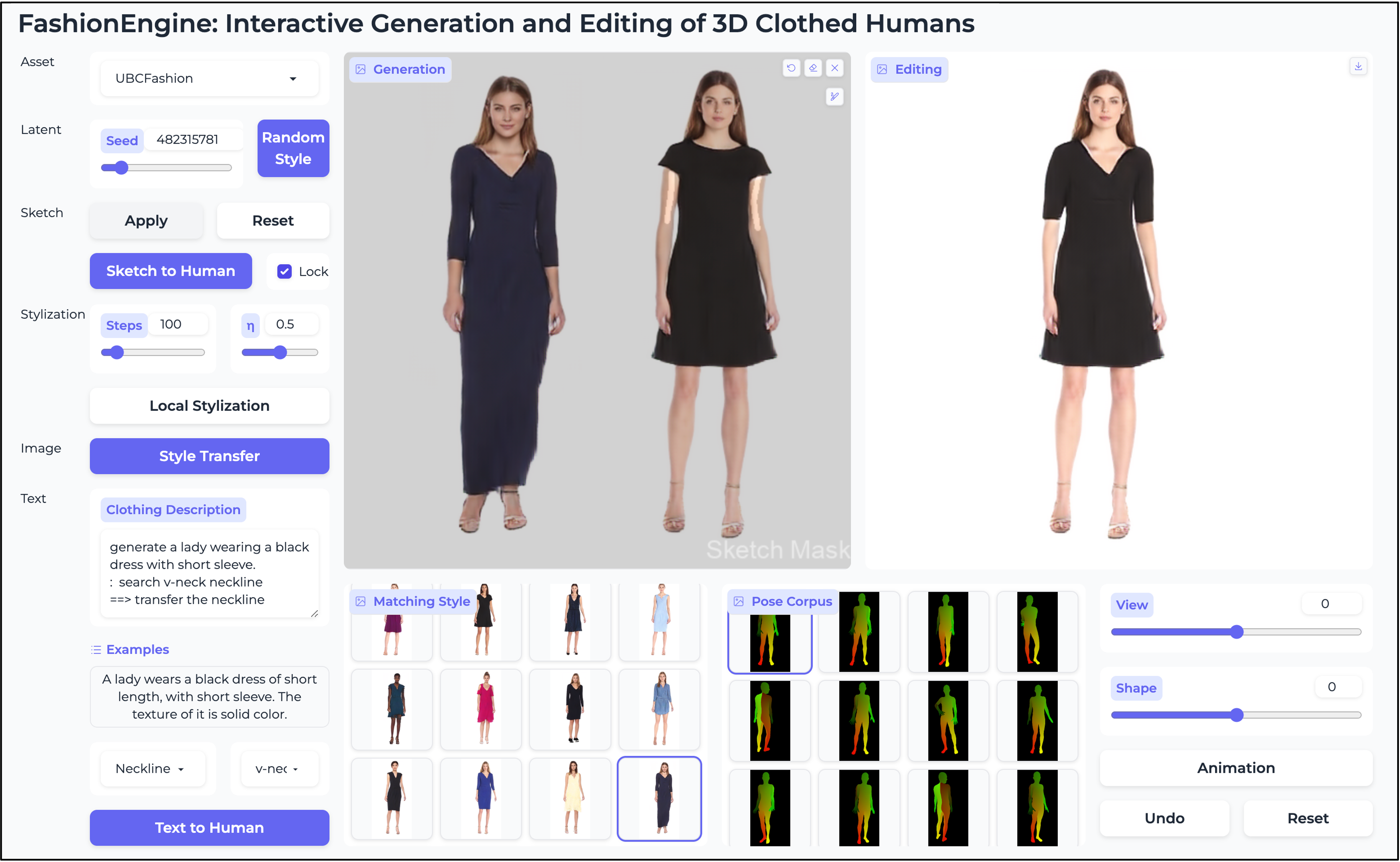}         
          \end{center}
	\vspace{-0.04in}
	\caption{\nickname{} User Interface (Sec \ref{sec:method_ui}). Users can generate humans by unconditional ('Random Style') or conditional generation (\eg, 'Text to Human'), and edit the generated humans by sketch-based ('Sketch to Human'), image-based ('Style Transfer'), and text-based ('Text-to-Human') editing. For conditional sketch or text input, \ie, text 'search v-neck neckline', 'Matching Style' is provided to search candidate styles (\eg, 'v-neck neckline' that match the input), and users are allowed to select desired styles for more flexible and generalizable style editing. Users can also check the generated humans under different poses by selecting one specific pose in 'Pose Corpus', and under different viewpoints or shapes by adjusting the 'View' or 'Shape' slider. The generated humans are also animatable. See the live demo for more details.}
	\label{fig:ui}
\end{figure*}
}

\newcommand{\figMMEdit}{
\begin{figure*}[]
	\begin{center}
		\includegraphics[width=\textwidth]{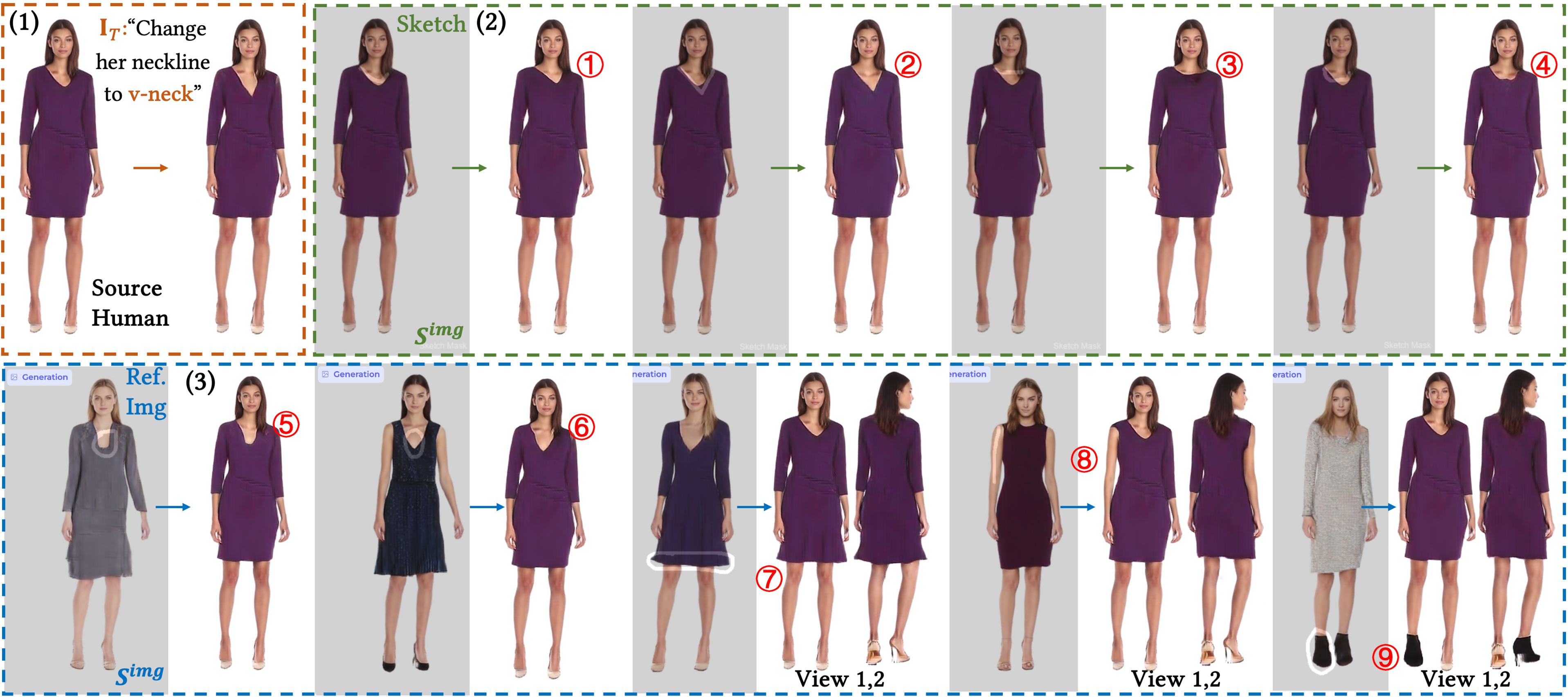}         
  \end{center}
	\vspace{-0.08in}
	\caption{The capacity of multimodal editing. Designers can simply input texts to edit humans (1), or draw sketches to describe the clothing shape for finer-grained editing (2), or transfer any parts of the clothing style from a reference image (3) for view-consistent editing.}
	\label{fig:mmedit}
\end{figure*}
}

\newcommand{\figAnyPose}{
\begin{figure}[]
	\begin{center}
		\includegraphics[width=.48\textwidth]{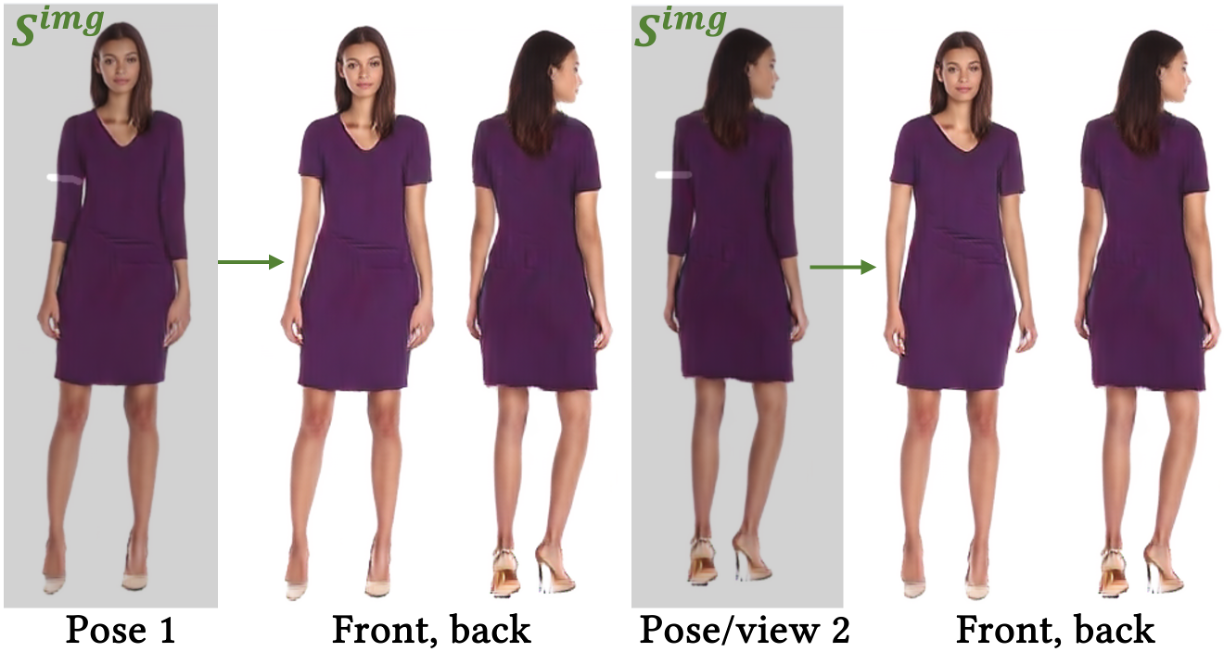}         
  \end{center}
	\vspace{-0.08in}
	\caption{Pose-, view-agnostic editing. \nickname{} allows designers to edit humans in different poses or viewpoints by transforming the input signals from the image space to the unified UV space.}
	\label{fig:anypose}
\end{figure}
}

\newcommand{\figSketchParser}{
\begin{figure}[t]
	\begin{center}
		\vspace{-0.08in}
		\includegraphics[width=.48\textwidth]{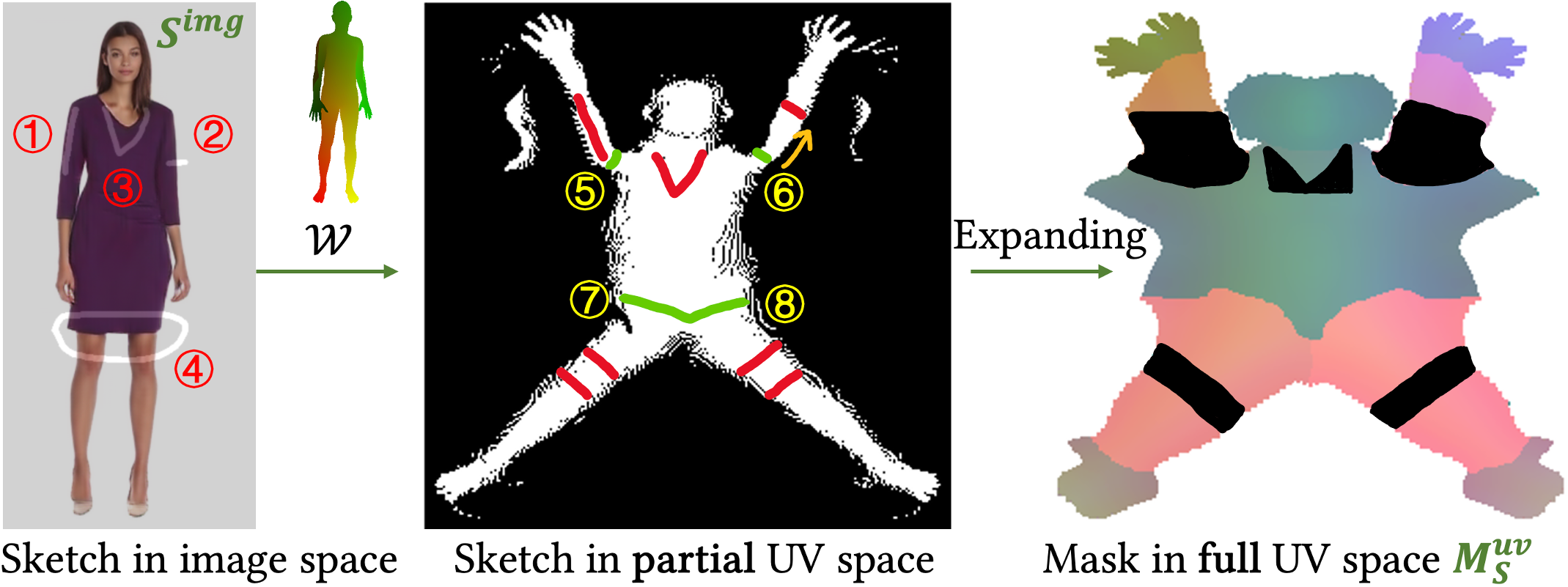}         
  \end{center}
	\caption{Illustration of Sketch Parser. Sketch Parser supports 4 different types of sketch input: sleeve or dress length in two manners \cnum{1}\cnum{2}, neckline shape \cnum{3}, or a closed area \cnum{4}. Sketches are transformed from the image space to the unified partial UV space (warped by single view as shown in Fig. \ref{fig:m2space}), and expanded into a mask in the full UV space based on body topology prior.}
	\label{fig:sketchparser}
\end{figure}
}

\newcommand{\figCmpDreampose}{
	\begin{figure}[]
		\begin{center}
			\includegraphics[width=.48\textwidth]{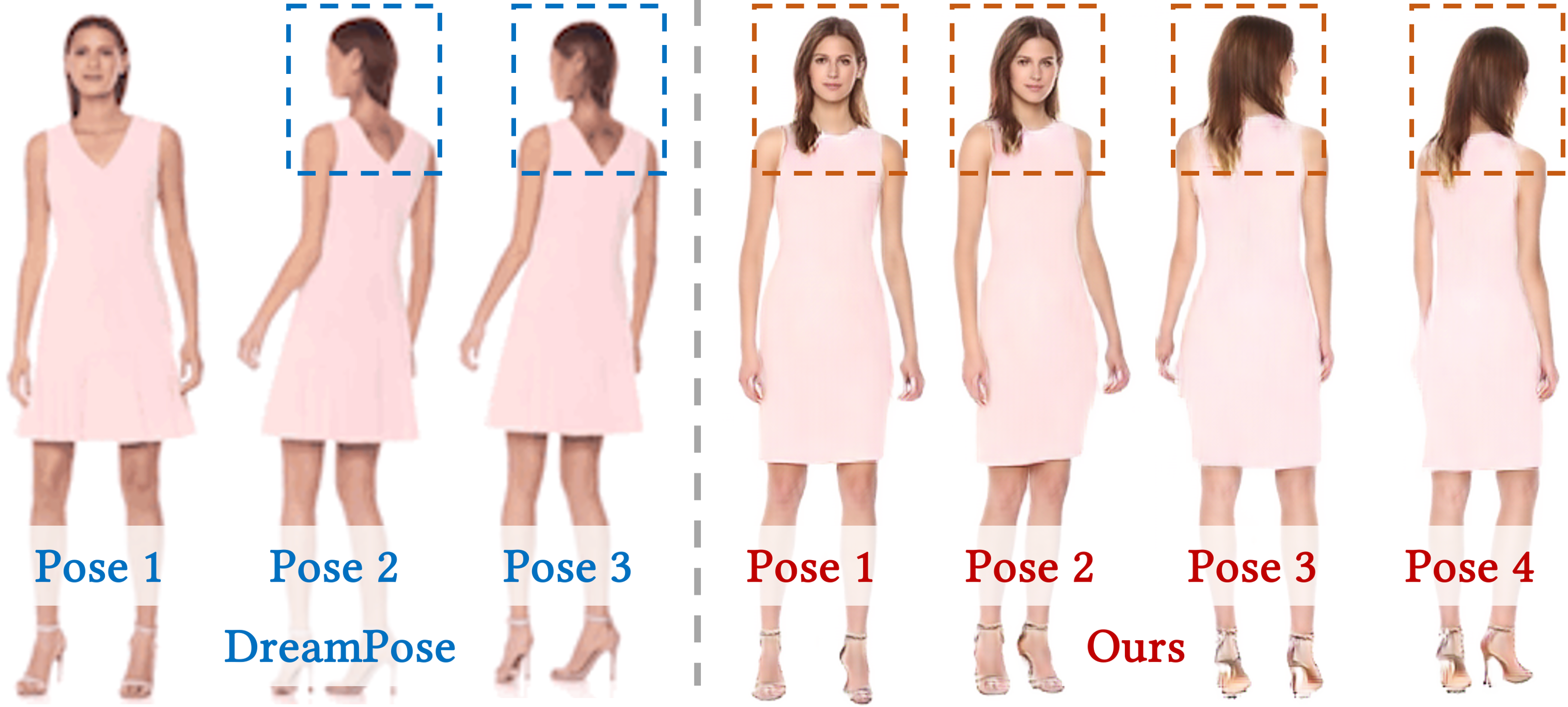}         
		\end{center}
		\vspace{-0.08in}
		\caption{Comparisons with DreamPose \cite{dreampose2023} for generations on the UBCFashion dataset. Ours synthesizes high-quality and consistent humans (e.g., hair) under different poses or viewpoints.}
		\label{fig:figCmpDreampose}
	\end{figure}
}

\newcommand{\figAbSmplShape}{
	\begin{figure}[]
		\begin{center}
			\includegraphics[width=.48\textwidth]{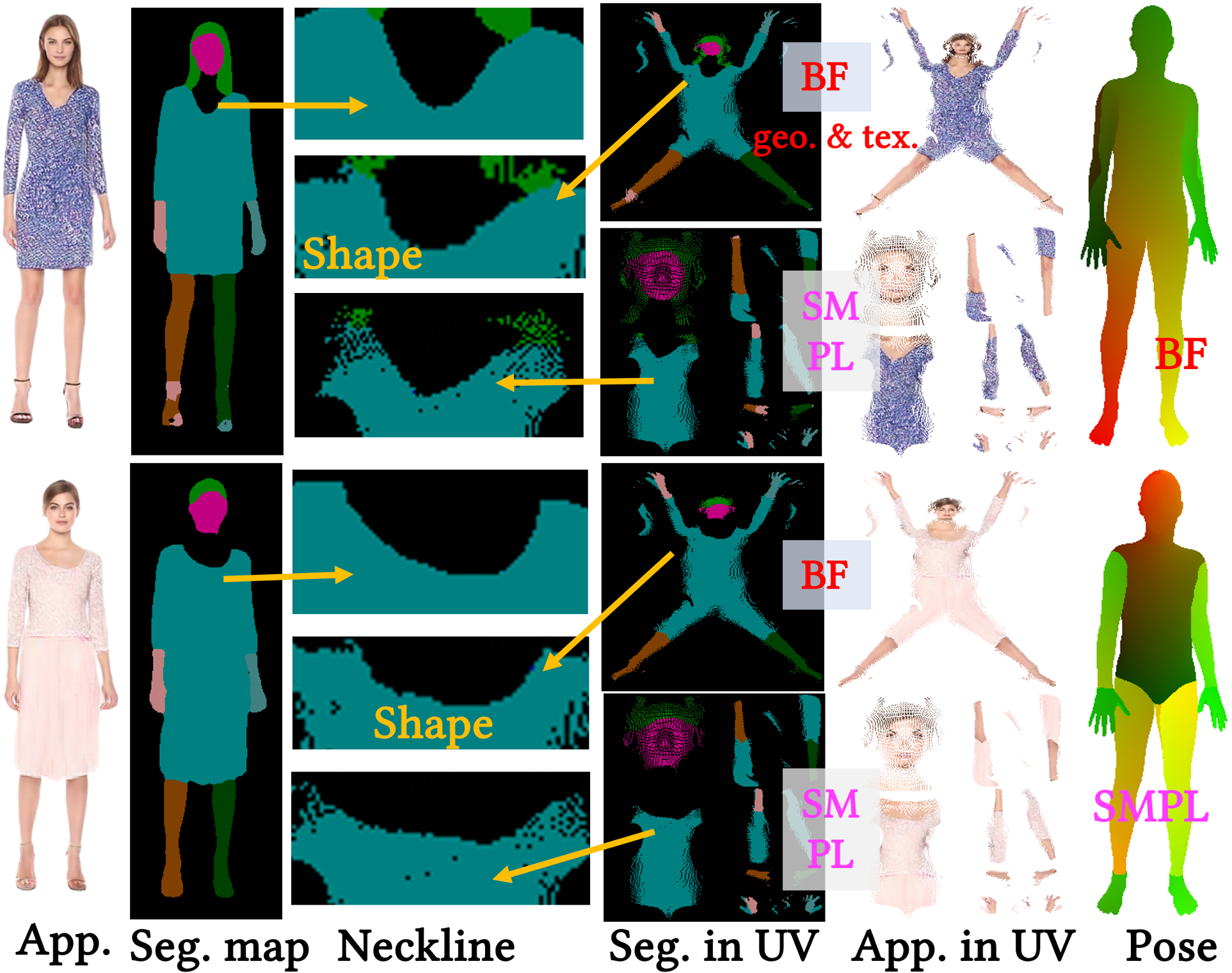}         
		\end{center}
		\vspace{-0.08in}
		\caption{Comparisons of boundary-free (BF) UV and standard SMPL UV. BF UV faithfully preserves the shape of neckline after UV warping, and preserves both the geometry and texture in a more readable way than the standard SMPL UV for editing.}
		\label{fig:figAbSmplShape}
	\end{figure}
}

\newcommand{\figAbFlling}{
	\begin{figure}[]
		\begin{center}
			\includegraphics[width=.48\textwidth]{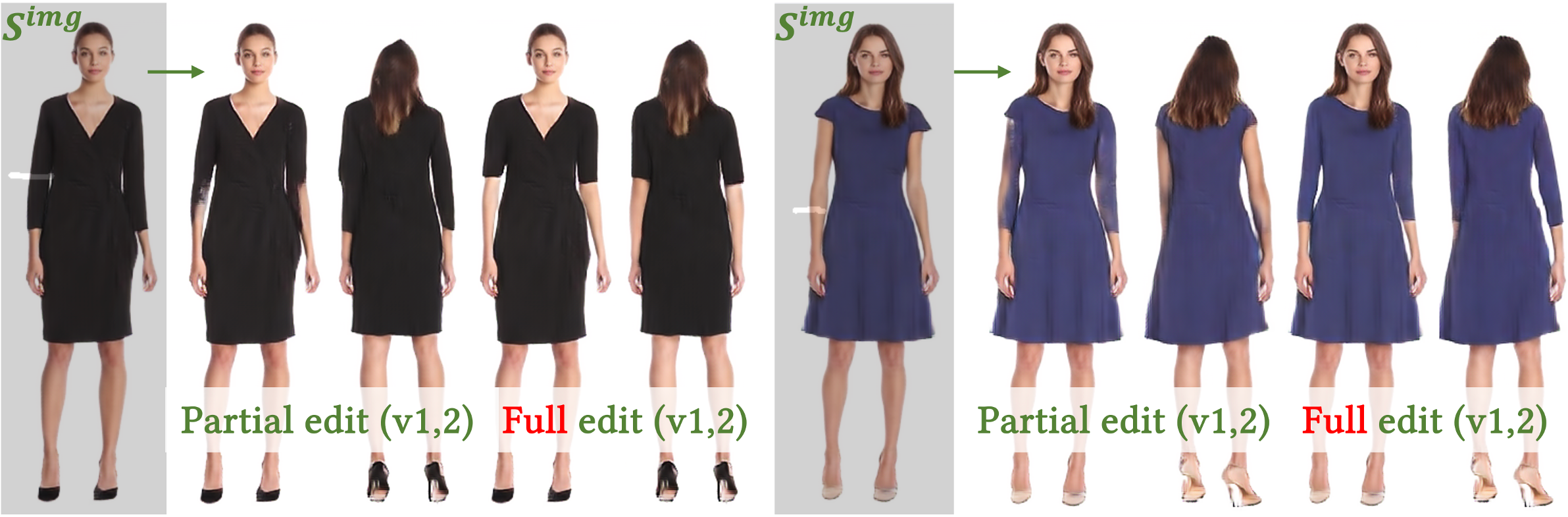}         
		\end{center}
		\caption{Ablation study of UV latent editing. Full UV latent editing (with the expansion step in Fig. \ref{fig:sketchparser}) enables view-consistent human renderings.}
		\label{fig:figAbFlling}
	\end{figure}
}

\begin{teaserfigure}
  \includegraphics[width=\textwidth]{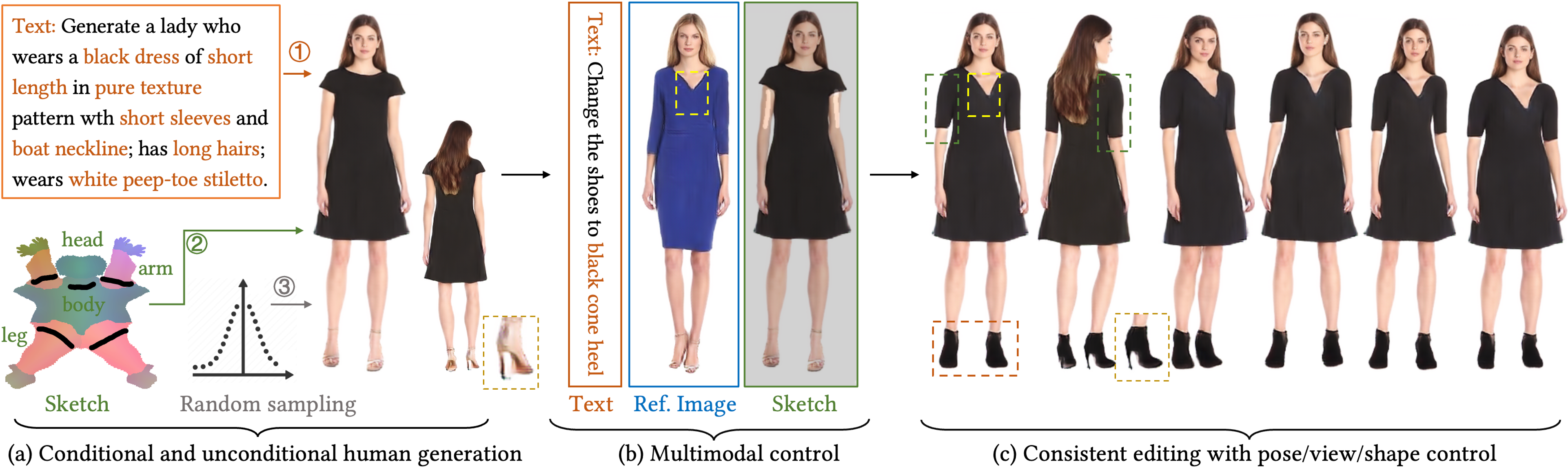}
  \caption{With \nickname{}, artist-designers can generate a view-consistent clothed human in three different ways (a), including \cnum{1} texts describing the human clothing; \cnum{2} hand-drawing sketches describing the clothing shape such as neckline shape, length of sleeve, and the length of lower clothing on a warped human body template; and \cnum{3} random appearance sampling. (b) Users are also allowed to edit the generated human interactively with multimodal control (e.g., texts, reference images, and sketches). (c) Users can adjust the pose and shape of the edited humans and check the renderings from different camera viewpoints before exporting images or video assets.}
  \label{fig:teaser}
\end{teaserfigure}

\begin{abstract}
We present \textbf{\nickname{}}, an interactive 3D human generation and editing system that creates 3D digital humans via user-friendly multimodal controls such as natural languages, visual perceptions, and hand-drawing sketches. \nickname{} automates the 3D human production with three key components: \textbf{1) A pre-trained 3D human diffusion model} that learns to model 3D humans in a semantic UV latent space from 2D image training data, which provides strong priors for diverse generation and editing tasks. \textbf{2) Multimodality-UV Space} encoding the texture appearance, shape topology, and textual semantics of human clothing in a canonical UV-aligned space, which faithfully aligns the user multimodal inputs with the implicit UV latent space for controllable 3D human editing. The multimodality-UV space is shared across different user inputs, such as texts, images, and sketches, which enables various joint multimodal editing tasks. \textbf{3) Multimodality-UV Aligned Sampler} learns to sample high-quality and diverse 3D humans from the diffusion prior. Extensive experiments validate \nickname's state-of-the-art performance for conditional generation/editing tasks. In addition, we present an interactive user interface for our FashionEngine that enables both conditional and unconditional generation tasks, and editing tasks including pose/view/shape control, text-, image-, and sketch-driven 3D human editing and 3D virtual try-on, in a unified framework. Our project page is at: \href{https://taohuumd.github.io/projects/FashionEngine/}{{\color{blue} https://taohuumd.github.io/projects/FashionEngine}}.
\end{abstract}

\ccsdesc[500]{Computing methodologies~Rendering}

%
%
\keywords{3D Human Generation, Editing}
\maketitle

\section{Introduction}

With the development of game, virtual reality and film industry, there is an increasing demand for high-quality 3D contents, especially 3D avatars. Traditionally, the production of 3D avatars requires days of work from highly skilled 3D content creators, which is not only time-consuming but also expensive. There have been some attempts in trying to automate the avatar generation pipeline~\cite{eva,ag3d}. However, they usually lack control over the generation process, making it difficult to use in practice, as shown in Tab.~\ref{tab:summary}. To reduce the friction of using learning-based avatar generation algorithms, we propose \nickname{}, an interactive system that enables the generation and editing of high-quality photo-realistic 3D humans. The process is controlled by multiple modalities, \eg, texts, images, and hand-drawing sketches, making the system easy to use even for layman users.

\nickname{} automates the 3D human production in three steps, as shown in Fig.~\ref{fig:teaser}. In the first step, a candidate 3D human is generated either randomly or conditionally from text descriptions or hand-drawing sketches. Subsequently, users can use text, reference images, or simply draw sketches to edit the appearance of the 3D human in an interactive way. In the last step, final adjustments in poses and shapes can be performed before being rendered into images or videos. A key challenge for human editing lies in the alignment between the user inputs and the human representation space. \nickname{} enables the human editing with three components.

Firstly, \nickname{} utilizes the 3D human priors learned in a pre-trained model \cite{structldm} that models humans in a semantic UV latent space. The UV latent space preserves the articulated structure of the human body and enables detailed appearance capture and editing. From the latent space, a 3D-aware auto-decoder is employed to embed the properties learned from the 2D training data, which decodes the latents into 3D humans under different poses, viewpoints, and clothing styles. Furthermore, a 3D diffusion model \cite{structldm} learns the latent space for generative human sampling, which serves as strong priors for different editing tasks.

Secondly, we construct a Multimodality-UV Space from the learned prior, which encodes the appearance, shape topology, and textual semantics of human clothing in a canonical UV space. The multimodal user inputs (e.g., texts, images, and sketches) are faithfully aligned with the implicit UV latent space for controllable 3D human editing. Notably, the multimodality-UV space is shared across different user inputs, which enables various joint multimodal editing tasks.

Thirdly, we propose Multimodality-UV Aligned Samplers conditioned on the user inputs to sample high-quality 3D humans from the diffusion prior. Specifically, a Text-UV Aligned Sampler and a Sketch-UV Aligned Sampler are proposed for text- or sketch-driven generation or editing, respectively. A key component is to search the Multimodality-UV Space to sample desired latents that are well aligned with the user inputs for controllable generation and editing.

Quantitative and qualitative experiments are performed on different generation and editing tasks, including conditional generation, and text-, image-, and sketch-driven 3D human editing. Experimental results illustrate the versatility and scalability of \nickname{}. In addition, our system runs at about 9.2 FPS to render $512^2$ resolution images on an NVIDIA V100 GPU, which enables interactive editing tasks. To conclude, our contributions are listed as follows.  

\textbf{1)} We propose an interactive 3D human generation and editing system, \nickname{}, which enables easy and efficient production of high-quality 3D humans.

\textbf{2)} Making use of a pre-trained 3D human prior, we propose a 3D human editing framework in the UV-latent space, which effectively unifies controlling signals from multiple modalities, \eg, texts, images, and sketches, for joint multimodal editing. 

\textbf{3)} Extensive experiments show the  state-of-the-art performances of the UV-based \nickname{} system in 3D human editing.

\section{Related Work}

\paragraph{Human Image Generation and Editing.} It has been a great success for single-category generation, \eg, human faces, with generative adversarial networks (GAN)~\cite{goodfellow2020generative, stylegan1, stylegan2, stylegan3}. However, generating the whole body with diverse clothing and poses is more challenging for GANs~\cite{humangan, tryongan, sarkar2021style, jiang2022text2human, stylepeople,Lin2023FashionTexCV,styleganhuman, fruhstuck2022insetgan}. The disentanglement of GAN latent space provides opportunities for image editing~\cite{zhu2016generative, voynov2020unsupervised, shen2021closed, harkonen2020ganspace, Wu2022DeepPortraitDrawingGH, Su2022DrawingInStylesPI}. DragGAN~\cite{pan2023drag} presents an intuitive interface for easy image editing. The recent success of diffusion models in general image generation has also motivated researchers to apply this method on human generation~\cite{controlnet, liu2023hyperhuman}.

\paragraph{3D Human Generation and Editing.} With the advancement of volume rendering~\cite{nerf}, some exiting approaches train 3D-aware generative networks from 2D image collections~\cite{pigan, giraffe, eg3d, hvtr,surmo}, and 3D-aware human GANs are studied that model garment and human in unified framework \cite{enarfgan,eva,ag3d,Xiong2023Get3DHumanLS}. These methods use 1D latent, which makes it challenging to perform editing. Other than the data-driven methods, 2D priors~\cite{clip, ldm} can also be used to enable text-to-3D human generation~\cite{avatarclip, cao2023dreamavatar, cao2023guide3d}. Most recently, 3D diffusion models are also employed for unconditional 3D human generations, such as PrimDiffusion~\cite{chen2023primdiffusion} and StructLDM \cite{structldm}. StructLDM models humans in boundary-free UV latent space \cite{Zeng20203DHM} in contrast to the tranditional UV space used in \cite{chen2023primdiffusion,egorend,hvtrpp}, and learns a latent diffusion model for unconditional generation. In this work, making use of UV-based latent space and the unconditional diffusion prior \cite{structldm}, we propose 3D human conditional editing with multimodal controls.

\tabSummary

\paragraph{3D Diffusion Models.} Diffusion models are proven to be able to capture complex distributions. Directly learning diffusion models on 3D representations has been explored in recent year. Diffusion models on point clouds~\cite{nichol2022point, luo2021diffusion, zeng2022lion}, voxels~\cite{zhou20213d, muller2022diffrf} and implicit models~\cite{jun2023shap} are used for coarse shape generation. As a compact 3D representation, tri-plane~\cite{eg3d} can be used with 2D network backbones for efficient 3D generation~\cite{shue20223d, wang2022rodin, gu2023nerfdiff, gupta20233dgen}.

\paragraph{Garment Modeling.}
Garments can also be separately modeled via 3D representations \cite{Qiu2023RECMVR3,
Zhu2020DeepFA,Chen2023ImplicitPCAIP,Zhu2022RegisteringET}. GarmentCode \cite{Korosteleva2023GarmentCodePP} designs synthetic parameterized garments. DressCode \cite{He2024DressCodeAS} learns text-driven 3D garment generations from synthetic 3D garments. \cite{Wolff2021DesigningPG} learns the garment shapes from 3D scans. In contrast, ours is built upon a generative model that is learned from real diverse internet videos, which enables photorealistic 3D human generation and editing with multimodal controls.
\section{FashionEngine}

\nickname{} learns to dress humans interactively and controllably, and it works in three stages. 1) In the first stage,  a 3D human appearance prior $\mathcal{Z}$ is learned from the training dataset by a latent diffusion model \cite{structldm} (Sec. \ref{sec:method_prior}). 2) With the prior $\mathcal{Z}$, users are allowed to generate a textured clothed 3D human by randomly sampling $\mathcal{Z}$ or uploading some texts describing the human appearance or a sketch mask describing the clothing mask for controllable generation (Sec. \ref{sec:method_generation}). 3) Users can optionally edit the generated humans by uploading the desired appearance styles in the form of texts, sketches, or reference images (Sec. \ref{sec:method_editing}).

\figMPrior

\subsection{3D Human Prior Learning}
\label{sec:method_prior}

\nickname{} is built upon StructLDM \cite{structldm} that models humans in a structured semantic UV-aligned space, and it learns 3D human priors in two stages. In the first stage, from a training dataset containing various human subject images with estimated SMPL and camera parameters distribution, an auto-decoder is learned with a set of structured embeddings ${Z}$ corresponding to the training human subjects. In the second stage, a latent diffusion model is learned from the embeddings ${Z}$ in the semantic UV latent space, which provides strong priors for diverse and realistic human generations. The pipeline of prior learning is depicted in Fig. \ref{fig:m1prior}, and refer to StructLDM \cite{structldm} for more details. We utilize the priors learned in StructLDM for the following editing tasks.

\subsection{Multimodality-UV Space}
With the human prior $\mathcal{Z}$ learned in the template UV space, we construct a multimodal-UV space that is aligned with the latent space for multimodal generation.

\label{sec:method_space} We sample a set of base latent $\mathbb{Z} \sim \mathcal{Z}$ from the learned prior, and construct a Multimodality-UV space by rendering and annotating the latent space. Each latent $z_i \sim \mathcal{Z}$ can be rendered into an image $A^{can}_i$ in a canonical space by renderer $\mathcal{R}$, under the fixed pose, shape, and camera intrinsic/extrinsic parameters. The rendered image $A^{can}_i$ is warped by $\mathcal{W}$ to UV space $A^{uv}_i$ using the UV correspondences (UV coordinate map in the posed space). Note that though $A^{uv}_i$ is warped from only a single partial view, it well preserves the clothing topology attributes in a user-readable way, such as the length of sleeves, neckline shape, and the length of lower clothing. In addition, each latent $z_i$ is annotated with detailed text descriptions as shown in Fig. \ref{fig:m2space}. We also render a segmentation map for each latent by \cite{li2020self}, which is warped to UV space ${Seg}^{uv}_i$. ${Seg}^{uv}_i$ preserves the clothing shape attributes such as the length of sleeves, neckline shape, and length of lower clothing. 

We render all the latents in $\mathbb{Z}$ to construct a multimodal space including an Appearance-Canonical Space (App-Can., $\mathbb{A}^{can}$), an Appearance-UV Space (App-UV., $\mathbb{A}^{uv}$), a textual Semantics-UV Space (Sem-UV, $\mathbb{T}^{uv}$), and Shape-UV Space ($\mathbb{S}^{uv}$) as shown in Fig. \ref{fig:m2space}.

\figMSpace
\figMGeneration

\subsection{Controllable Multimodal Generation}
\label{sec:method_generation}

\nickname{} generates human images given text input $\mathbf{I}_T$ or sketch input $\mathbf{I}^{uv}_S$ in the template UV space. We present UV-aligned samplers to sample the multimodal-UV space for controllable multimodal generation (Sec. \ref{sec:method_textgen}, \ref{sec:method_sketchgen}).

\subsubsection{Text-Driven Generation.} \label{sec:method_textgen}

Users are allowed to upload input texts $\mathbb{I}_T$ describing the appearances of the desired human, including hairstyle, and six clothing properties: clothing color, texture pattern, neckline shape, sleeve length, length of lower clothing, and types of shoe, as shown in Fig. \ref{fig:m2gen}, which semantically corresponds to six parts $P$=\{head, neck, body, arm, leg, foot\} in the UV space. A Text Parser $\mathcal{P}_T$ (Sec. \ref{sec:sketchparser}) is proposed to align the input texts with semantic body parts in UV space, which yields part-aware Text-UV aligned semantics $\mathbf{I}^{uv}_T = \mathcal{P}_T(\mathbf{I}_T)$.

\paragraph{Text-UV Aligned Sampler.} We further propose a Text-UV Aligned Sampler $\mathcal{T}$ to sample an latent $z^{*}_{T}$ from the learned prior $\mathcal{Z}$ conditioned on the input text $z^{*}_{T} = \mathcal{T}(\mathbf{I}^{uv}_T; \mathcal{Z})$. The sampler works in two stages. In the first stage, we independently search the latents in the Sem-UV Space $\mathbb{T}^{uv}$ that match the input text descriptions $\mathbf{I}^{uv}_T$ semantically for each body part $p \in P$ through \textit{SemMatch}: 
\begin{equation}
\label{eq:sem_match}   
   Z_p = \underset{z_i \in \mathbb{Z}, t^{uv}_i \in \mathbb{T}^{uv}}{\arg\max} SemMatch(t^{uv}_i[p], \mathbf{I}^{uv}_T[p])
\end{equation}
where $\mathbf{I}^{uv}_T[p]$ indicates the extraction of the textual semantics of part $p$. In the paper, we consider top-n matches, and E.q. \ref{eq:sem_match} yields a set of candidate latents for each body part $p$, $Z_p = \{z_0, ..., z_{k-1}\}$.

\label{sec:appMatch}
In the second step, we find the best match latent $z^*_p$ for each part based on the appearance similarity score by \textit{AppMatch}:
\begin{equation}
\label{eq:m_gen_app_match}
   \max_{z^*_p \in Z_p, A^{uv}_p \subset \mathbb{A}^{uv}} AppMatch(\{A^{uv}_p[M_{body}] | p \in P\})
\end{equation}
where $A^{uv}_p = \{A^{uv}_i | z_i \in Z_p\}$, and $M_{body}$ indicates the body mask in UV space  (\cnum{1} in Fig. \ref{fig:m2space}(b)) in UV space. In the paper, we calculate the multichannel SSIM \cite{ssim} score in \textit{AppMatch}.

With $\{z^*_p \in Z_p\}$, an optimal latent $z^{*}_{T}$ is constructed conditioned on input text ${\mathbf{I}_T}$:
\begin{equation}
\label{eq:m_gen_construct}
   z^{*}_{T} = \sum_{ p \in P} z^*_{p} * M_p
\end{equation}

\noindent where $M_p$ is the mask of body part $p$. 


To explain this process, let's take Fig. \ref{fig:m2gen} as an example. For the sake of clarity, we will only consider two attributes: hairstyle and sleeve length. For the input text \{"long hair", "long sleeve"\}, we get $\{ Z_{head}=\{z_k\}, Z_{arm}=\{z_i, z_m\} \}$  where $\{z_k\}$ meets "long hair", and $\{z_i, z_m\}$ meets "long sleeve", as shown in Fig. \ref{fig:m2gen}(b). The group $\{z*_{head}=z_k, z*_{arm} = z_i\}$ is ranked as the optimal since  $AppMatch(\cnum{1}, \cnum{2})$ $>$ $AppMatch(\cnum{1}, \cnum{4})$ in terms of SSIM score. Note that instead of calculating the similarity in image space, we warp the patches (e.g., \cnum{1} in Fig. \ref{fig:m2gen}) to a compact UV space (e.g., \cnum{1} Fig. \ref{fig:m2space}) for efficiency, and it also eliminates the effects of the tightness of different clothing types in evaluation. We finally get $z^{*}_{T} = z_k * M_{head} + z_i * M_{arm}$, which is rendered as images by Diff-Render, $\mathbf{Y}_T = \mathcal{R} \circ \mathcal{D} (z^{*}_{T})$.

\subsubsection{Sketch-Driven Generation.} \label{sec:method_sketchgen} Users are also allowed to design a human by simply sketching to describe the dress shape including neckline shape, length of sleeve, and lower clothing in the template UV space, $\mathbf{I}^{uv}_S$ as shown in Fig. \ref{fig:m2gen} (a). A Sketch Parser (Sec. \ref{sec:sketchparser}) is employed to translate the raw sketch into a clothing mask $M^{uv}_S$.  

\paragraph{Sketch-UV Aligned Sampler.} We further propose a Sketch-UV Aligned Sampler $\mathcal{S}$ to sample a latent $z^{*}_{T} = \mathcal{T}(I^{uv}_M; \mathcal{Z})$ from the learned prior $\mathcal{Z}$ conditioned on the sketch mask $M^{uv}_S$ in two stages. In the first stage, we search the latents in the Shape-UV Space $\mathbb{S}^{uv}$ that matches the input sketch mask $M^{uv}_S$ for body part $p$ using \textit{ShapeMatch}:
\begin{equation}
\label{eq:sketch_match}   
   Z_p = \underset{z_i \in \mathbb{Z}, s^{uv}_i \in \mathbb{S}^{uv}}{\arg\min} ShapeMatch(s^{uv}_i[p], M^{uv}_S[p])
\end{equation}
where $M^{uv}_S[p]$ indicates the pixels of part $p$. The \textit{ShapeMatch} algorithm evaluates the shape similarity between two binary masks, and it is the lower the result, the better match it is. It is calculated based on the hu-moment values \cite{shapematch}. In the paper, we consider top-k matches, and E.q. \ref{eq:sketch_match} yields a set of candidate latents for each body part $p$. We employ \textit{AppMatch} E.q. \ref{eq:m_gen_app_match} to find the optimal latent $\{z^*_p | p \in P\}$, and target latent $z^*_S$ is similarly constructed via E.q. \ref{eq:m_gen_construct}.

We take Fig. \ref{fig:m2gen} (a) as an example to explain the process. For the sake of clarity, only two attributes are considered: neckline shape and sleeve length. For the input clothing mask $M^{uv}_S$ which is parsed as "a dress shape with medium length of lower clothing, sleeves cut off, and V-shape neckline", we get the matched latents $\{Z_{neck}=\{z_i, z_m\}, Z_{arm}=\{z_j\}\}$ as shown in Fig. \ref{fig:m2gen}(b).

The group $\{z^*_{neck}=z_i, z^*_{arm} = z_j\}$ is ranked as the optimal since $AppMatch(\cnum{3}, \cnum{2})$ $>$ $AppMatch(\cnum{3}, \cnum{4})$ in terms of SSIM score. We finally get $z^{*}_S = z_i * M_{neck} + z_j * M_{arm}$, which is rendered as images by Diff-Render, $\mathbf{Y}_S = \mathcal{R} \circ \mathcal{D} (z^{*}_S)$.

\figMEdit

\subsection{Controllable Multimodal Editing}
\label{sec:method_editing}
\nickname{} also allows users to edit the generated humans by uploading texts, sketches, or guided images to describe the desired clothing appearances, as shown in Fig. \ref{fig:m3edit}.

\subsubsection{Text-Driven Editing.} \label{sec:textedit} For the generated human with source latent $z$, given input text $\mathcal{I}_T$ describing the editing commands, the Text Parser $\mathcal{P}_T$ is employed to parse the input in UV space, which yields $\mathcal{I}^{uv}_T$ and a mask $M^{uv}_T$ indicating which body part to edit. The Text-UV aligned sampler $\mathcal{T}$ searches the Sem-UV Space $\mathbb{T}^{uv}$ to find an optimal latent $z^*_T$ that matches the input editing texts by \textit{SemMatch} E.q. \ref{eq:sem_match} with a higher appearance similarity by \textit{AppMatch} E.q. \ref{eq:m_gen_app_match}:  $z^{*}_T = \mathcal{T}(\mathbf{I}^{uv}_T, z; \mathcal{Z})$. Note that a preprocessing step is to render $z$ to the canonical space as introduced in Sec. \ref{sec:method_space}, yielding a warped image $A^{uv}$ in UV space for appearance matching in \textit{AppMatch}. Finally, the updated latent is calculated as $z'_T = z^*_T * M^{uv}_T + (1 - M^{uv}_T) * z$, which can be rendered by Diff-Render \cite{structldm}. 

Fig. \ref{fig:m3edit} shows the qualitative results, which suggest that given text descriptions, \nickname{} is capable of synthesizing view-consistent appearance editing results for sleeves (1), neckline (2), and shoes (3).

\figSketchParser

\subsubsection{Sketch-Driven Editing.} \label{sec:sketchedit} Users can also edit the clothing style by simply drawing sketches for finer-grained control. For a generated human image with source latent \textit{z} and input sketches in image space $S^{img}$, users can extend the sleeves \cnum{1}\cnum{3}, change the neckline to V-shape \cnum{2}, and extend the length of lower clothing \cnum{4} as shown in Fig. \ref{fig:m3edit}. We first transform the input sketches from image space to UV space by $\mathcal{W}$, and a Sketch Parser $\mathcal{P}_S$ (Sec. \ref{sec:sketchparser}) is employed to synthesize clothing masks $M^{uv}_S$ for editing. Conditioned on the source latent $z$ and mask, the Sketch-UV aligned sampler $\mathcal{S}$ searches the Shape-UV Space $\mathbb{S}^{uv}$ to find an optimal latent $z^*_S$ that matches the sketch mask by \textit{ShapeMatch} E.q. \ref{eq:sketch_match} with a higher appearance similarity by \textit{AppMatch} E.q. \ref{eq:m_gen_app_match}:  $z^{*}_S = \mathcal{S}(\mathbf{M}^{uv}_T, z; \mathcal{Z})$. Similar to text-driven editing, we render a canonical image of $z$ for \textit{AppMatch}. Finally, the updated latent is calculated as $z'_S = z^*_S * M^{uv}_S + (1 - M^{uv}_S) * z$, which can be rendered by Diff-Render. Fig. \ref{fig:m3edit} shows the consistent clothing rendering that is well aligned with the input sketch, including V-shape neckline \cnum{4}, short left and right sleeves \cnum{5}\cnum{6}, and dress \cnum{7}.

\subsubsection{Image-Driven Editing.} \label{sec:imgedit} A picture is worth a thousand words; texts sometimes struggle to describe a specific clothing style, whereas images provide concrete references. Given a reference image of a dressed human, \nickname{} allows users to transfer any parts of the clothing style from the reference image to edit the source humans by \textit{joint image- and sketch-driven editing}, as shown in Fig. \ref{fig:m3edit}. Users can sketch to mark which parts of clothing style will be transferred, e.g., \cnum{5}\cnum{6}\cnum{7}\cnum{8} $S^{img}$ shown in Fig. \ref{fig:m3edit}. 

We transform the sketch from image space to UV space by $\mathcal{W}$, and utilize a Sketch Parser $\mathcal{P}_S$ to synthesize clothing masks $M^{uv}_I$ for editing. Given the reference latent $z_R$, the source latent $z$ is updated to $z'_I = z_R * M^{uv}_I + z * (1 - M^{uv}_I)$., which is rendered into images by Diff-Render. Fig. \ref{fig:m3edit} shows that selected styles are faithfully transferred to the source human including the style of right arm \cnum{7}, hemline structure \cnum{8}, neckline \cnum{9} and shoes.

\figUI

\subsection{Interactive User Interface}
\label{sec:method_ui}
We present an interactive user interface for our \nickname{} as shown in Fig. \ref{fig:ui}. \nickname{} provides users with  unconditional and conditional human generations, and three different manners of sketch-, image-, and text-based editing. Adjustments in pose/view/shape are also allowed, and users can generate and export animation videos using \nickname{}. Refer to the \textbf{live demo} for more details.

\subsection{Method Appendix: Text and Sketch Parser}
\paragraph{Text Parser} Due to the scale of dataset, instead of learning a language model to encode the text inputs, the Text Parser uses keyword matching as used in Text2Human \cite{jiang2022text2human} based on a similar template \cite{jiang2023text2performer}.

\label{sec:sketchparser}
\paragraph{Sketch Parser} An illustration is shown in Fig. \ref{fig:sketchparser}, where 4 different types of sketch input are allowed to design the clothing style, including the length of sleeve or dress by \cnum{1} or \cnum{2}, the shape of neckline \cnum{3}, and a closed area of clothing \cnum{4}. Sketches in the image space are transformed to the unified partial UV space as shown in Fig. \ref{fig:m2space}. Note that the partial UV space preserves the body topology, e.g., the boundary of each body part \cnum{5}\cnum{6}\cnum{7}\cnum{8}, which allows to expand the partial sketches into clothing masks in the full UV space that is aligned with the latent space for faithful editing.

\section{Experiments}

\figedit

\subsection{Experimental Setup}

\paragraph{Datasets and Metrics} We perform experiments on a monocular video dataset UBCFashion~\cite{ubcfashion} that contains 500 monocular human videos with natural fashion motions. We conduct a perceptual user study and quantitative comparisons to evaluate the visual quality, consistency with input, and identity preservation ability in different human editing tasks.

\subsection{Comparisons to State-of-the-Art Methods}

\paragraph{Text-Driven Editing.} We compare against InstructPix2Pix for text-driven human image editing, as shown in Fig. \ref{fig:edit} (a). We generate high-quality images, and the editing results are well-aligned with the text inputs. In addition, we also faithfully preserve the identity information, whereas InstructPix2Pix cannot achieve this. The advantages over InstructPix2Pix are further confirmed by the user study in Fig. \ref{fig:userstudy}, where 25 participants are asked to select the images with better visual quality, better consistency with the text inputs, and better preservation of identity information. It suggests that more than 90\% of editing images by our methods are considered to be more realistic than InstructPix2Pix.

Quantitatively, we compute the CLIP~\cite{clip} feature similarity, dubbed CLIP score, between 20 randomly generated pairs of editing results and text prompts to show the effectiveness of the text-driven editing. As shown in Tab.~\ref{tab:clipscore}, CLIP score of our results consistently outperforms that of InstructPix2Pix, which shows the stronger instruction-following ability of our method. We also analyze the identity preservation ability by evaluating the face similarity with the source images, which is defined as the cosine distance of FaceNet \cite{Schroff2015FaceNetAU} embeddings of face images. We also evaluate the appearance preservation ability after local editing by measuring the PSNR on unaffected areas. Tab.~\ref{tab:clipscore} suggests both the strong identity and appearance preservation ability of our method. 


\paragraph{Sketch-Driven Editing.} We also compare against DragGAN \cite{pan2023drag} for sketch-driven human image editing, as shown in Fig. \ref{fig:edit} (b). Our method supports local editing, and we faithfully preserve the identity information. However, DragGAN often edits humans in a global latent space and the local editing generally affects the full-body appearances, and hence the identity information is not well preserved. A user study in Fig. \ref{fig:userstudy} (b) further confirms this, where 25 participants are asked to select the images with better visual quality, better consistency with the sketch inputs, and better preservation of identity information. More than 85\% participants prefer our results in terms of visual quality, and about 80\% participants prefer our method for consistency with sketch input and identity preservation.

\paragraph{3D vs. 2D Human Diffusion.}
We compare against the 2D diffusion approach DreamPose \cite{dreampose2023} for generations on novel poses, as shown in Fig. \ref{fig:figCmpDreampose}. DreamPose fine-tunes a  pre-trained Stable Diffusion module on the UBCFashion dataset for human generations. We select the generated humans with similar appearances for comparisons. Fig. \ref{fig:figCmpDreampose} shows the advantages of the 3D diffusion approach over DreamPose in consistent human renderings (e.g., hair) under different poses or viewpoints. In addition, our system synthesizes higher-quality images than DreamPose.

\figUserStudy

\tabclipscore

\figAb

\subsection{Ablation Study}

\tabAbMixer
\paragraph{Receptive Field (RF) in Global Style Mixer.} We utilize the architecture of StructLDM \cite{structldm}, where a global style mixer is employed to learn full-body appearance style. We evaluate how the size of RF affects the reconstruction quality by comparing the performances of reconstructing humans under RF=2 and RF=4 on about 4,000 images in the autodecoding stage. The qualitative results are shown in Fig. \ref{fig:ab} (b), which suggests that a bigger Receptive Field (RF=4) captures the global clothing styles better than RF=2 \cnum{3}, and also recovers more details than RF=2, such as the high heels \cnum{4}. In addition, RF=4 successfully reconstructs the hemline and between-leg offsets of the dress, while a smaller RF fails \cnum{5}. The conclusion is further confirmed by the quantitative comparisons listed in Tab. \ref{tab:AbMixer}, where a bigger RF achieves better quantitative results on LPIPS \cite{lpip}, FID \cite{Heusel2017GANsTB} and PSNR. Note that  the global style mixer also upsamples features in style mixing, and RF=2 upsamples the original 2D feature maps from $256^2$ to $512^2$ image resolutions, while RF=4 upsamples the features from $128^2$ to $512^2$.

\figMMEdit

\figAbSmplShape

\paragraph{Appearance Match.} As presented in Sec. \ref{sec:appMatch}, an Appearance Match (\textit{AppMatch}) technique is required to sample desired latents for both Text-UV Aligned Sampler and Sketch-UV Aligned Sampler. For the cases in Fig. \ref{fig:m2gen}, we provide more intermediate details to analyze the effects of the \textit{AppMatch}, as shown in Fig. \ref{fig:ab} (a). For both the text-driven and sketch-driven generation tasks in Fig. \ref{fig:m2gen}, we get some candidate latents by TextMatch or ShapeMatch, such as Cand.1, Cand.2, and Cand.3, which all meet the requirements of "long sleeves" and "V-shape neckline". With \textit{AppMatch}, the sleeves of Cand. 1 will be transferred to the source identity 1 (Src1) since \textit{AppMatch} (Src1, Cand. 1) $>$ \textit{AppMatch} (Src1, Cand. 2), which enables higher quality generation than Cand. 2 or 3. Similarly, \textit{AppMatch} also improves the results for sketch-driven generation tasks, such as the generation of V-shape neckline \cnum{2}.

\paragraph{Boundary-Free UV vs. Standard SMPL UV.} We employ a continuous boundary-free (BF) UV topology \cite{Zeng20203DHM} for editing, and the comparisons against standard SMPL UV are shown in Fig. \ref{fig:figAbSmplShape}. BF preserves both the geometry and texture in a more readable way than the standard SMPL UV, which offers user-friendly interfaces for sketch-driven generation as shown in Fig. \ref{fig:teaser}, \ref{fig:m2gen}. In addition, BF UV faithfully preserves the clothing shape after UV warping, \ie the neckline shape in BF UV has more similarities to the source image than the standard SMPL UV. The BF UV mapping has also been proven to be more friendly for CNN-based architecture than the standard discontinuous UV mapping in \cite{Zeng20203DHM}.

\figAbFlling

\paragraph{Partial vs. Full UV Latent Editing.} We also compare the differences of editing humans in the partial or full UV space, as illustrated in Fig. \ref{fig:figAbFlling}. The main difference lies in whether the expansion step is included as shown in Fig. \ref{fig:sketchparser}. Partial editing is similar to image-based editing, which generates inconsistent humans, whereas our full UV latent editing enables view-consistent human renderings.


\figAnyPose

\subsection{Controllable 3D Human Editing}

\paragraph{Multimodal Editing.} Fig. \ref{fig:mmedit} illustrates each modality (texts, sketches, and image-based style transfer) in human editing. Texts are the {easiest} way to edit humans, \ie "changing the neckline to v-neck", whereas they sometimes fails to efficiently and accurately convey the designer's intent, such as the size and orientation of the v-neck. To solve this issue, sketch-driven editing is provided for \textit{finer-grained} editing, \ie, precisely controlling the size of the v-neck \cnum{1}\cnum{2}, boat/round neckline \cnum{3}\cnum{4} by simply drawing sketches.   

A picture is worth a thousand words. We improve the \textit{capacity of editing} by providing image-based style transfer as introduced in Sec. \ref{sec:imgedit}. Designers can freely transfer any parts of the clothing style from the reference image to edit the source humans \ie, transferring the neckline \cnum{5}\cnum{6}, wavy hem \cnum{7}, sleeve style \cnum{8} and shoes \cnum{9}. Note that to get the reference styles/images, designers can utilize the functionality of \textit{'Matching Style'} that searches the candidate styles based on rough text or sketch descriptions. Refer to Fig. \ref{fig:ui} and the live demo for more details.

\paragraph{Pose-View-Shape Control.} \nickname{} also integrates the functionality of pose/viewpoint/shape control for human rendering, which are achieved by changing the camera or human template parameters (\eg, SMPL). Refer to the live demo for the functionality. 

\paragraph{Pose-, View-Agnostic Editing.} \nickname{} allows artist-designers to edit humans in different poses and viewpoints, which is achieved by transforming the input signals (e.g., sketches) from the image space to the unified UV space (illustrated in Fig. \ref{fig:m3edit}, \ref{fig:sketchparser}) for pose- and view-independent editing as shown in Fig. \ref{fig:anypose}.

\section{Discussion}

We propose \nickname{}, an interactive 3D human generation and editing system, which allows easy and efficient production of 3D digital human with multimodal control. \nickname{} models the multimodal features in a unified UV space that enables various joint multimodal editing tasks. We illustrate the advantages of our UV-based editing system in different 3D human editing tasks. 

\paragraph{Limitations.} Our synthesized humans are biased toward generating females with dresses due to the dataset bias, same as \cite{jiang2023text2performer,structldm,dreampose2023}. More data can be involved in the training to alleviate this bias. Our system is capable of editing challenging skirts, which shows potentials for other clothing styles. 

\figCmpDreampose

\section*{Acknowledgement}

This study is supported by the Ministry of Education, Singapore, under its MOE AcRF Tier 2 (MOET2EP20221- 0012), NTU NAP, and under the RIE2020 Industry Alignment Fund – Industry Collaboration Projects (IAF-ICP) Funding Initiative, as well as cash and in-kind contribution from the industry partner(s).

\bibliographystyle{ACM-Reference-Format}
\bibliography{main_short}


\begin{thebibliography}{69}


\ifx \showCODEN    \undefined \def \showCODEN     #1{\unskip}     \fi
\ifx \showDOI      \undefined \def \showDOI       #1{#1}\fi
\ifx \showISBNx    \undefined \def \showISBNx     #1{\unskip}     \fi
\ifx \showISBNxiii \undefined \def \showISBNxiii  #1{\unskip}     \fi
\ifx \showISSN     \undefined \def \showISSN      #1{\unskip}     \fi
\ifx \showLCCN     \undefined \def \showLCCN      #1{\unskip}     \fi
\ifx \shownote     \undefined \def \shownote      #1{#1}          \fi
\ifx \showarticletitle \undefined \def \showarticletitle #1{#1}   \fi
\ifx \showURL      \undefined \def \showURL       {\relax}        \fi
\providecommand\bibfield[2]{#2}
\providecommand\bibinfo[2]{#2}
\providecommand\natexlab[1]{#1}
\providecommand\showeprint[2][]{arXiv:#2}

\bibitem[Brooks et~al\mbox{.}(2023)]%
        {brooks2022instructpix2pix}
\bibfield{author}{\bibinfo{person}{Tim Brooks}, \bibinfo{person}{Aleksander Holynski}, {and} \bibinfo{person}{Alexei~A. Efros}.} \bibinfo{year}{2023}\natexlab{}.
\newblock \showarticletitle{InstructPix2Pix: Learning to Follow Image Editing Instructions}. In \bibinfo{booktitle}{\emph{CVPR}}.
\newblock


\bibitem[Cao et~al\mbox{.}(2023a)]%
        {cao2023dreamavatar}
\bibfield{author}{\bibinfo{person}{Yukang Cao}, \bibinfo{person}{Yan-Pei Cao}, \bibinfo{person}{Kai Han}, \bibinfo{person}{Ying Shan}, {and} \bibinfo{person}{Kwan-Yee~K Wong}.} \bibinfo{year}{2023}\natexlab{a}.
\newblock \showarticletitle{Dreamavatar: Text-and-shape guided 3d human avatar generation via diffusion models}.
\newblock \bibinfo{journal}{\emph{arXiv preprint arXiv:2304.00916}} (\bibinfo{year}{2023}).
\newblock


\bibitem[Cao et~al\mbox{.}(2023b)]%
        {cao2023guide3d}
\bibfield{author}{\bibinfo{person}{Yukang Cao}, \bibinfo{person}{Yan-Pei Cao}, \bibinfo{person}{Kai Han}, \bibinfo{person}{Ying Shan}, {and} \bibinfo{person}{Kwan-Yee~K Wong}.} \bibinfo{year}{2023}\natexlab{b}.
\newblock \showarticletitle{Guide3D: Create 3D Avatars from Text and Image Guidance}.
\newblock \bibinfo{journal}{\emph{arXiv preprint arXiv:2308.09705}} (\bibinfo{year}{2023}).
\newblock


\bibitem[Chan et~al\mbox{.}(2022)]%
        {eg3d}
\bibfield{author}{\bibinfo{person}{Eric~R Chan}, \bibinfo{person}{Connor~Z Lin}, \bibinfo{person}{Matthew~A Chan}, \bibinfo{person}{Koki Nagano}, \bibinfo{person}{Boxiao Pan}, \bibinfo{person}{Shalini De~Mello}, \bibinfo{person}{Orazio Gallo}, \bibinfo{person}{Leonidas~J Guibas}, \bibinfo{person}{Jonathan Tremblay}, \bibinfo{person}{Sameh Khamis}, {et~al\mbox{.}}} \bibinfo{year}{2022}\natexlab{}.
\newblock \showarticletitle{Efficient geometry-aware 3D generative adversarial networks}. In \bibinfo{booktitle}{\emph{Proceedings of the IEEE/CVF Conference on Computer Vision and Pattern Recognition}}. \bibinfo{pages}{16123--16133}.
\newblock


\bibitem[Chan et~al\mbox{.}(2021)]%
        {pigan}
\bibfield{author}{\bibinfo{person}{Eric~R Chan}, \bibinfo{person}{Marco Monteiro}, \bibinfo{person}{Petr Kellnhofer}, \bibinfo{person}{Jiajun Wu}, {and} \bibinfo{person}{Gordon Wetzstein}.} \bibinfo{year}{2021}\natexlab{}.
\newblock \showarticletitle{pi-gan: Periodic implicit generative adversarial networks for 3d-aware image synthesis}. In \bibinfo{booktitle}{\emph{Proceedings of the IEEE/CVF conference on computer vision and pattern recognition}}. \bibinfo{pages}{5799--5809}.
\newblock


\bibitem[Chen et~al\mbox{.}(2023b)]%
        {Chen2023ImplicitPCAIP}
\bibfield{author}{\bibinfo{person}{Lan Chen}, \bibinfo{person}{Jie Yang}, \bibinfo{person}{Hongbo Fu}, \bibinfo{person}{Xiaoxu Meng}, \bibinfo{person}{Weikai Chen}, \bibinfo{person}{Bo Yang}, {and} \bibinfo{person}{Lin Gao}.} \bibinfo{year}{2023}\natexlab{b}.
\newblock \showarticletitle{ImplicitPCA: Implicitly-proxied parametric encoding for collision-aware garment reconstruction}.
\newblock \bibinfo{journal}{\emph{Graph. Model.}}  \bibinfo{volume}{129} (\bibinfo{year}{2023}), \bibinfo{pages}{101195}.
\newblock
\urldef\tempurl%
\url{https://api.semanticscholar.org/CorpusID:262155303}
\showURL{%
\tempurl}


\bibitem[Chen et~al\mbox{.}(2023a)]%
        {chen2023primdiffusion}
\bibfield{author}{\bibinfo{person}{Zhaoxi Chen}, \bibinfo{person}{Fangzhou Hong}, \bibinfo{person}{Haiyi Mei}, \bibinfo{person}{Guangcong Wang}, \bibinfo{person}{Lei Yang}, {and} \bibinfo{person}{Ziwei Liu}.} \bibinfo{year}{2023}\natexlab{a}.
\newblock \showarticletitle{PrimDiffusion: Volumetric Primitives Diffusion for 3D Human Generation}. In \bibinfo{booktitle}{\emph{Thirty-seventh Conference on Neural Information Processing Systems}}.
\newblock


\bibitem[Dong et~al\mbox{.}(2023)]%
        {ag3d}
\bibfield{author}{\bibinfo{person}{Zijian Dong}, \bibinfo{person}{Xu Chen}, \bibinfo{person}{Jinlong Yang}, \bibinfo{person}{Michael~J. Black}, \bibinfo{person}{Otmar Hilliges}, {and} \bibinfo{person}{Andreas Geiger}.} \bibinfo{year}{2023}\natexlab{}.
\newblock \showarticletitle{AG3D: Learning to Generate 3D Avatars from 2D Image Collections}.
\newblock \bibinfo{journal}{\emph{ArXiv}}  \bibinfo{volume}{abs/2305.02312} (\bibinfo{year}{2023}).
\newblock
\urldef\tempurl%
\url{https://api.semanticscholar.org/CorpusID:258461509}
\showURL{%
\tempurl}


\bibitem[Fr{\"u}hst{\"u}ck et~al\mbox{.}(2022)]%
        {fruhstuck2022insetgan}
\bibfield{author}{\bibinfo{person}{Anna Fr{\"u}hst{\"u}ck}, \bibinfo{person}{Krishna~Kumar Singh}, \bibinfo{person}{Eli Shechtman}, \bibinfo{person}{Niloy~J Mitra}, \bibinfo{person}{Peter Wonka}, {and} \bibinfo{person}{Jingwan Lu}.} \bibinfo{year}{2022}\natexlab{}.
\newblock \showarticletitle{InsetGAN for Full-Body Image Generation}. In \bibinfo{booktitle}{\emph{Proceedings of the IEEE/CVF Conference on Computer Vision and Pattern Recognition}}. \bibinfo{pages}{7723--7732}.
\newblock


\bibitem[Fu et~al\mbox{.}(2022)]%
        {styleganhuman}
\bibfield{author}{\bibinfo{person}{Jianglin Fu}, \bibinfo{person}{Shikai Li}, \bibinfo{person}{Yuming Jiang}, \bibinfo{person}{Kwan-Yee Lin}, \bibinfo{person}{Chen Qian}, \bibinfo{person}{Chen~Change Loy}, \bibinfo{person}{Wayne Wu}, {and} \bibinfo{person}{Ziwei Liu}.} \bibinfo{year}{2022}\natexlab{}.
\newblock \showarticletitle{StyleGAN-Human: A Data-Centric Odyssey of Human Generation}. In \bibinfo{booktitle}{\emph{European Conference on Computer Vision}}.
\newblock
\urldef\tempurl%
\url{https://api.semanticscholar.org/CorpusID:248377018}
\showURL{%
\tempurl}


\bibitem[Goodfellow et~al\mbox{.}(2020)]%
        {goodfellow2020generative}
\bibfield{author}{\bibinfo{person}{Ian Goodfellow}, \bibinfo{person}{Jean Pouget-Abadie}, \bibinfo{person}{Mehdi Mirza}, \bibinfo{person}{Bing Xu}, \bibinfo{person}{David Warde-Farley}, \bibinfo{person}{Sherjil Ozair}, \bibinfo{person}{Aaron Courville}, {and} \bibinfo{person}{Yoshua Bengio}.} \bibinfo{year}{2020}\natexlab{}.
\newblock \showarticletitle{Generative adversarial networks}.
\newblock \bibinfo{journal}{\emph{Commun. ACM}} \bibinfo{volume}{63}, \bibinfo{number}{11} (\bibinfo{year}{2020}), \bibinfo{pages}{139--144}.
\newblock


\bibitem[Grigorev et~al\mbox{.}(2021)]%
        {stylepeople}
\bibfield{author}{\bibinfo{person}{Artur Grigorev}, \bibinfo{person}{Karim Iskakov}, \bibinfo{person}{Anastasia Ianina}, \bibinfo{person}{Renat Bashirov}, \bibinfo{person}{Ilya Zakharkin}, \bibinfo{person}{Alexander Vakhitov}, {and} \bibinfo{person}{Victor~S. Lempitsky}.} \bibinfo{year}{2021}\natexlab{}.
\newblock \showarticletitle{StylePeople: A Generative Model of Fullbody Human Avatars}.
\newblock \bibinfo{journal}{\emph{2021 (CVPR)}} (\bibinfo{year}{2021}), \bibinfo{pages}{5147--5156}.
\newblock


\bibitem[Gu et~al\mbox{.}(2023)]%
        {gu2023nerfdiff}
\bibfield{author}{\bibinfo{person}{Jiatao Gu}, \bibinfo{person}{Alex Trevithick}, \bibinfo{person}{Kai-En Lin}, \bibinfo{person}{Joshua~M Susskind}, \bibinfo{person}{Christian Theobalt}, \bibinfo{person}{Lingjie Liu}, {and} \bibinfo{person}{Ravi Ramamoorthi}.} \bibinfo{year}{2023}\natexlab{}.
\newblock \showarticletitle{Nerfdiff: Single-image view synthesis with nerf-guided distillation from 3d-aware diffusion}. In \bibinfo{booktitle}{\emph{International Conference on Machine Learning}}. PMLR, \bibinfo{pages}{11808--11826}.
\newblock


\bibitem[Gupta et~al\mbox{.}(2023)]%
        {gupta20233dgen}
\bibfield{author}{\bibinfo{person}{Anchit Gupta}, \bibinfo{person}{Wenhan Xiong}, \bibinfo{person}{Yixin Nie}, \bibinfo{person}{Ian Jones}, {and} \bibinfo{person}{Barlas O{\u{g}}uz}.} \bibinfo{year}{2023}\natexlab{}.
\newblock \showarticletitle{3dgen: Triplane latent diffusion for textured mesh generation}.
\newblock \bibinfo{journal}{\emph{arXiv preprint arXiv:2303.05371}} (\bibinfo{year}{2023}).
\newblock


\bibitem[H{\"a}rk{\"o}nen et~al\mbox{.}(2020)]%
        {harkonen2020ganspace}
\bibfield{author}{\bibinfo{person}{Erik H{\"a}rk{\"o}nen}, \bibinfo{person}{Aaron Hertzmann}, \bibinfo{person}{Jaakko Lehtinen}, {and} \bibinfo{person}{Sylvain Paris}.} \bibinfo{year}{2020}\natexlab{}.
\newblock \showarticletitle{Ganspace: Discovering interpretable gan controls}.
\newblock \bibinfo{journal}{\emph{Advances in neural information processing systems}}  \bibinfo{volume}{33} (\bibinfo{year}{2020}), \bibinfo{pages}{9841--9850}.
\newblock


\bibitem[He et~al\mbox{.}(2024)]%
        {He2024DressCodeAS}
\bibfield{author}{\bibinfo{person}{Kai He}, \bibinfo{person}{Kaixin Yao}, \bibinfo{person}{Qixuan Zhang}, \bibinfo{person}{Jingyi Yu}, \bibinfo{person}{Lingjie Liu}, {and} \bibinfo{person}{Lan Xu}.} \bibinfo{year}{2024}\natexlab{}.
\newblock \showarticletitle{DressCode: Autoregressively Sewing and Generating Garments from Text Guidance}.
\newblock \bibinfo{journal}{\emph{ArXiv}}  \bibinfo{volume}{abs/2401.16465} (\bibinfo{year}{2024}).
\newblock
\urldef\tempurl%
\url{https://api.semanticscholar.org/CorpusID:267320968}
\showURL{%
\tempurl}


\bibitem[Heusel et~al\mbox{.}(2017)]%
        {Heusel2017GANsTB}
\bibfield{author}{\bibinfo{person}{Martin Heusel}, \bibinfo{person}{Hubert Ramsauer}, \bibinfo{person}{Thomas Unterthiner}, \bibinfo{person}{Bernhard Nessler}, {and} \bibinfo{person}{Sepp Hochreiter}.} \bibinfo{year}{2017}\natexlab{}.
\newblock \showarticletitle{GANs Trained by a Two Time-Scale Update Rule Converge to a Local Nash Equilibrium}. In \bibinfo{booktitle}{\emph{NIPS}}.
\newblock


\bibitem[Hong et~al\mbox{.}(2022a)]%
        {eva}
\bibfield{author}{\bibinfo{person}{Fangzhou Hong}, \bibinfo{person}{Zhaoxi Chen}, \bibinfo{person}{Yushi Lan}, \bibinfo{person}{Liang Pan}, {and} \bibinfo{person}{Ziwei Liu}.} \bibinfo{year}{2022}\natexlab{a}.
\newblock \showarticletitle{EVA3D: Compositional 3D Human Generation from 2D Image Collections}.
\newblock \bibinfo{journal}{\emph{ArXiv}}  \bibinfo{volume}{abs/2210.04888} (\bibinfo{year}{2022}).
\newblock
\urldef\tempurl%
\url{https://api.semanticscholar.org/CorpusID:252780848}
\showURL{%
\tempurl}


\bibitem[Hong et~al\mbox{.}(2022b)]%
        {avatarclip}
\bibfield{author}{\bibinfo{person}{Fangzhou Hong}, \bibinfo{person}{Mingyuan Zhang}, \bibinfo{person}{Liang Pan}, \bibinfo{person}{Zhongang Cai}, \bibinfo{person}{Lei Yang}, {and} \bibinfo{person}{Ziwei Liu}.} \bibinfo{year}{2022}\natexlab{b}.
\newblock \showarticletitle{Avatarclip: Zero-shot text-driven generation and animation of 3d avatars}.
\newblock \bibinfo{journal}{\emph{arXiv preprint arXiv:2205.08535}} (\bibinfo{year}{2022}).
\newblock


\bibitem[Hu(1962)]%
        {shapematch}
\bibfield{author}{\bibinfo{person}{Ming-Kuei Hu}.} \bibinfo{year}{1962}\natexlab{}.
\newblock \showarticletitle{Visual pattern recognition by moment invariants}.
\newblock \bibinfo{journal}{\emph{IRE Transactions on Information Theory}} \bibinfo{volume}{8}, \bibinfo{number}{2} (\bibinfo{year}{1962}), \bibinfo{pages}{179--187}.
\newblock
\urldef\tempurl%
\url{https://doi.org/10.1109/TIT.1962.1057692}
\showDOI{\tempurl}


\bibitem[Hu et~al\mbox{.}(2024a)]%
        {structldm}
\bibfield{author}{\bibinfo{person}{Tao Hu}, \bibinfo{person}{Fangzhou Hong}, {and} \bibinfo{person}{Ziwei Liu}.} \bibinfo{year}{2024}\natexlab{a}.
\newblock \bibinfo{title}{StructLDM: Structured Latent Diffusion for 3D Human Generation}.
\newblock
\newblock
\showeprint[arxiv]{2404.01241}~[cs.CV]


\bibitem[Hu et~al\mbox{.}(2024b)]%
        {surmo}
\bibfield{author}{\bibinfo{person}{Tao Hu}, \bibinfo{person}{Fangzhou Hong}, {and} \bibinfo{person}{Ziwei Liu}.} \bibinfo{year}{2024}\natexlab{b}.
\newblock \bibinfo{title}{SurMo: Surface-based 4D Motion Modeling for Dynamic Human Rendering}.
\newblock
\newblock
\showeprint[arxiv]{2404.01225}~[cs.CV]


\bibitem[Hu et~al\mbox{.}(2021)]%
        {egorend}
\bibfield{author}{\bibinfo{person}{Tao Hu}, \bibinfo{person}{Kripasindhu Sarkar}, \bibinfo{person}{Lingjie Liu}, \bibinfo{person}{Matthias Zwicker}, {and} \bibinfo{person}{Christian Theobalt}.} \bibinfo{year}{2021}\natexlab{}.
\newblock \showarticletitle{EgoRenderer: Rendering Human Avatars From Egocentric Camera Images}. In \bibinfo{booktitle}{\emph{ICCV}}.
\newblock


\bibitem[Hu et~al\mbox{.}(2023)]%
        {hvtrpp}
\bibfield{author}{\bibinfo{person}{Tao Hu}, \bibinfo{person}{Hongyi Xu}, \bibinfo{person}{Linjie Luo}, \bibinfo{person}{Tao Yu}, \bibinfo{person}{Zerong Zheng}, \bibinfo{person}{He Zhang}, \bibinfo{person}{Yebin Liu}, {and} \bibinfo{person}{Matthias Zwicker}.} \bibinfo{year}{2023}\natexlab{}.
\newblock \showarticletitle{HVTR++: Image and Pose Driven Human Avatars using Hybrid Volumetric-Textural Rendering}.
\newblock \bibinfo{journal}{\emph{IEEE Transactions on Visualization and Computer Graphics}} (\bibinfo{year}{2023}), \bibinfo{pages}{1--15}.
\newblock
\urldef\tempurl%
\url{https://doi.org/10.1109/TVCG.2023.3297721}
\showDOI{\tempurl}


\bibitem[Hu et~al\mbox{.}(2022)]%
        {hvtr}
\bibfield{author}{\bibinfo{person}{Tao Hu}, \bibinfo{person}{Tao Yu}, \bibinfo{person}{Zerong Zheng}, \bibinfo{person}{He Zhang}, \bibinfo{person}{Yebin Liu}, {and} \bibinfo{person}{Matthias Zwicker}.} \bibinfo{year}{2022}\natexlab{}.
\newblock \showarticletitle{HVTR: Hybrid Volumetric-Textural Rendering for Human Avatars}.
\newblock \bibinfo{journal}{\emph{3DV}} (\bibinfo{year}{2022}).
\newblock


\bibitem[Jiang et~al\mbox{.}(2023)]%
        {jiang2023text2performer}
\bibfield{author}{\bibinfo{person}{Yuming Jiang}, \bibinfo{person}{Shuai Yang}, \bibinfo{person}{Tong~Liang Koh}, \bibinfo{person}{Wayne Wu}, \bibinfo{person}{Chen~Change Loy}, {and} \bibinfo{person}{Ziwei Liu}.} \bibinfo{year}{2023}\natexlab{}.
\newblock \showarticletitle{Text2Performer: Text-Driven Human Video Generation}. In \bibinfo{booktitle}{\emph{Proceedings of the IEEE/CVF International Conference on Computer Vision}}.
\newblock


\bibitem[Jiang et~al\mbox{.}(2022)]%
        {jiang2022text2human}
\bibfield{author}{\bibinfo{person}{Yuming Jiang}, \bibinfo{person}{Shuai Yang}, \bibinfo{person}{Haonan Qiu}, \bibinfo{person}{Wayne Wu}, \bibinfo{person}{Chen~Change Loy}, {and} \bibinfo{person}{Ziwei Liu}.} \bibinfo{year}{2022}\natexlab{}.
\newblock \showarticletitle{Text2Human: Text-Driven Controllable Human Image Generation}.
\newblock \bibinfo{journal}{\emph{ACM Transactions on Graphics (TOG)}} \bibinfo{volume}{41}, \bibinfo{number}{4}, Article \bibinfo{articleno}{162} (\bibinfo{year}{2022}), \bibinfo{numpages}{11}~pages.
\newblock
\urldef\tempurl%
\url{https://doi.org/10.1145/3528223.3530104}
\showDOI{\tempurl}


\bibitem[Jun and Nichol(2023)]%
        {jun2023shap}
\bibfield{author}{\bibinfo{person}{Heewoo Jun} {and} \bibinfo{person}{Alex Nichol}.} \bibinfo{year}{2023}\natexlab{}.
\newblock \showarticletitle{Shap-e: Generating conditional 3d implicit functions}.
\newblock \bibinfo{journal}{\emph{arXiv preprint arXiv:2305.02463}} (\bibinfo{year}{2023}).
\newblock


\bibitem[Karras et~al\mbox{.}(2023)]%
        {dreampose2023}
\bibfield{author}{\bibinfo{person}{Johanna~Suvi Karras}, \bibinfo{person}{Aleksander Holynski}, \bibinfo{person}{Ting-Chun Wang}, {and} \bibinfo{person}{Ira Kemelmacher-Shlizerman}.} \bibinfo{year}{2023}\natexlab{}.
\newblock \showarticletitle{DreamPose: Fashion Image-to-Video Synthesis via Stable Diffusion}.
\newblock \bibinfo{journal}{\emph{2023 IEEE/CVF International Conference on Computer Vision (ICCV)}} (\bibinfo{year}{2023}), \bibinfo{pages}{22623--22633}.
\newblock


\bibitem[Karras et~al\mbox{.}(2021)]%
        {stylegan3}
\bibfield{author}{\bibinfo{person}{Tero Karras}, \bibinfo{person}{Miika Aittala}, \bibinfo{person}{Samuli Laine}, \bibinfo{person}{Erik H\"ark\"onen}, \bibinfo{person}{Janne Hellsten}, \bibinfo{person}{Jaakko Lehtinen}, {and} \bibinfo{person}{Timo Aila}.} \bibinfo{year}{2021}\natexlab{}.
\newblock \showarticletitle{Alias-Free Generative Adversarial Networks}. In \bibinfo{booktitle}{\emph{Proc. NeurIPS}}.
\newblock


\bibitem[Karras et~al\mbox{.}(2019)]%
        {stylegan1}
\bibfield{author}{\bibinfo{person}{Tero Karras}, \bibinfo{person}{Samuli Laine}, {and} \bibinfo{person}{Timo Aila}.} \bibinfo{year}{2019}\natexlab{}.
\newblock \showarticletitle{A style-based generator architecture for generative adversarial networks}. In \bibinfo{booktitle}{\emph{Proceedings of the IEEE/CVF conference on computer vision and pattern recognition}}. \bibinfo{pages}{4401--4410}.
\newblock


\bibitem[Karras et~al\mbox{.}(2020)]%
        {stylegan2}
\bibfield{author}{\bibinfo{person}{Tero Karras}, \bibinfo{person}{Samuli Laine}, \bibinfo{person}{Miika Aittala}, \bibinfo{person}{Janne Hellsten}, \bibinfo{person}{Jaakko Lehtinen}, {and} \bibinfo{person}{Timo Aila}.} \bibinfo{year}{2020}\natexlab{}.
\newblock \showarticletitle{Analyzing and improving the image quality of stylegan}. In \bibinfo{booktitle}{\emph{Proceedings of the IEEE/CVF conference on computer vision and pattern recognition}}. \bibinfo{pages}{8110--8119}.
\newblock


\bibitem[Korosteleva and Sorkine-Hornung(2023)]%
        {Korosteleva2023GarmentCodePP}
\bibfield{author}{\bibinfo{person}{Maria Korosteleva} {and} \bibinfo{person}{Olga Sorkine-Hornung}.} \bibinfo{year}{2023}\natexlab{}.
\newblock \showarticletitle{GarmentCode: Programming Parametric Sewing Patterns}.
\newblock \bibinfo{journal}{\emph{ACM Transactions on Graphics (TOG)}}  \bibinfo{volume}{42} (\bibinfo{year}{2023}), \bibinfo{pages}{1 -- 15}.
\newblock
\urldef\tempurl%
\url{https://api.semanticscholar.org/CorpusID:259089061}
\showURL{%
\tempurl}


\bibitem[Lewis et~al\mbox{.}(2021)]%
        {tryongan}
\bibfield{author}{\bibinfo{person}{Kathleen~M Lewis}, \bibinfo{person}{Srivatsan Varadharajan}, {and} \bibinfo{person}{Ira Kemelmacher-Shlizerman}.} \bibinfo{year}{2021}\natexlab{}.
\newblock \showarticletitle{Tryongan: Body-aware try-on via layered interpolation}.
\newblock \bibinfo{journal}{\emph{ACM Transactions on Graphics (TOG)}} \bibinfo{volume}{40}, \bibinfo{number}{4} (\bibinfo{year}{2021}), \bibinfo{pages}{1--10}.
\newblock


\bibitem[Li et~al\mbox{.}(2020)]%
        {li2020self}
\bibfield{author}{\bibinfo{person}{Peike Li}, \bibinfo{person}{Yunqiu Xu}, \bibinfo{person}{Yunchao Wei}, {and} \bibinfo{person}{Yi Yang}.} \bibinfo{year}{2020}\natexlab{}.
\newblock \showarticletitle{Self-Correction for Human Parsing}.
\newblock \bibinfo{journal}{\emph{IEEE Transactions on Pattern Analysis and Machine Intelligence}} (\bibinfo{year}{2020}).
\newblock
\urldef\tempurl%
\url{https://doi.org/10.1109/TPAMI.2020.3048039}
\showDOI{\tempurl}


\bibitem[Lin et~al\mbox{.}(2023)]%
        {Lin2023FashionTexCV}
\bibfield{author}{\bibinfo{person}{Anran Lin}, \bibinfo{person}{Nanxuan Zhao}, \bibinfo{person}{Shuliang Ning}, \bibinfo{person}{Yuda Qiu}, \bibinfo{person}{Baoyuan Wang}, {and} \bibinfo{person}{Xiaoguang Han}.} \bibinfo{year}{2023}\natexlab{}.
\newblock \showarticletitle{FashionTex: Controllable Virtual Try-on with Text and Texture}.
\newblock \bibinfo{journal}{\emph{ACM SIGGRAPH 2023 Conference Proceedings}} (\bibinfo{year}{2023}).
\newblock
\urldef\tempurl%
\url{https://api.semanticscholar.org/CorpusID:258556972}
\showURL{%
\tempurl}


\bibitem[Liu et~al\mbox{.}(2023)]%
        {liu2023hyperhuman}
\bibfield{author}{\bibinfo{person}{Xian Liu}, \bibinfo{person}{Jian Ren}, \bibinfo{person}{Aliaksandr Siarohin}, \bibinfo{person}{Ivan Skorokhodov}, \bibinfo{person}{Yanyu Li}, \bibinfo{person}{Dahua Lin}, \bibinfo{person}{Xihui Liu}, \bibinfo{person}{Ziwei Liu}, {and} \bibinfo{person}{Sergey Tulyakov}.} \bibinfo{year}{2023}\natexlab{}.
\newblock \showarticletitle{HyperHuman: Hyper-Realistic Human Generation with Latent Structural Diffusion}.
\newblock \bibinfo{journal}{\emph{arXiv preprint arXiv:2310.08579}} (\bibinfo{year}{2023}).
\newblock


\bibitem[Luo and Hu(2021)]%
        {luo2021diffusion}
\bibfield{author}{\bibinfo{person}{Shitong Luo} {and} \bibinfo{person}{Wei Hu}.} \bibinfo{year}{2021}\natexlab{}.
\newblock \showarticletitle{Diffusion probabilistic models for 3d point cloud generation}. In \bibinfo{booktitle}{\emph{Proceedings of the IEEE/CVF Conference on Computer Vision and Pattern Recognition}}. \bibinfo{pages}{2837--2845}.
\newblock


\bibitem[Mildenhall et~al\mbox{.}(2020)]%
        {nerf}
\bibfield{author}{\bibinfo{person}{Ben Mildenhall}, \bibinfo{person}{Pratul~P. Srinivasan}, \bibinfo{person}{Matthew Tancik}, \bibinfo{person}{Jonathan~T. Barron}, \bibinfo{person}{Ravi Ramamoorthi}, {and} \bibinfo{person}{Ren Ng}.} \bibinfo{year}{2020}\natexlab{}.
\newblock \showarticletitle{NeRF: Representing Scenes as Neural Radiance Fields for View Synthesis}. In \bibinfo{booktitle}{\emph{ECCV}}.
\newblock


\bibitem[M{\"u}ller et~al\mbox{.}(2022)]%
        {muller2022diffrf}
\bibfield{author}{\bibinfo{person}{Norman M{\"u}ller}, \bibinfo{person}{Yawar Siddiqui}, \bibinfo{person}{Lorenzo Porzi}, \bibinfo{person}{Samuel~Rota Bul{\`o}}, \bibinfo{person}{Peter Kontschieder}, {and} \bibinfo{person}{Matthias Nie{\ss}ner}.} \bibinfo{year}{2022}\natexlab{}.
\newblock \showarticletitle{DiffRF: Rendering-Guided 3D Radiance Field Diffusion}.
\newblock \bibinfo{journal}{\emph{arXiv preprint arXiv:2212.01206}} (\bibinfo{year}{2022}).
\newblock


\bibitem[Nichol et~al\mbox{.}(2022)]%
        {nichol2022point}
\bibfield{author}{\bibinfo{person}{Alex Nichol}, \bibinfo{person}{Heewoo Jun}, \bibinfo{person}{Prafulla Dhariwal}, \bibinfo{person}{Pamela Mishkin}, {and} \bibinfo{person}{Mark Chen}.} \bibinfo{year}{2022}\natexlab{}.
\newblock \showarticletitle{Point-E: A System for Generating 3D Point Clouds from Complex Prompts}.
\newblock \bibinfo{journal}{\emph{arXiv preprint arXiv:2212.08751}} (\bibinfo{year}{2022}).
\newblock


\bibitem[Niemeyer and Geiger(2021)]%
        {giraffe}
\bibfield{author}{\bibinfo{person}{Michael Niemeyer} {and} \bibinfo{person}{Andreas Geiger}.} \bibinfo{year}{2021}\natexlab{}.
\newblock \showarticletitle{Giraffe: Representing scenes as compositional generative neural feature fields}. In \bibinfo{booktitle}{\emph{Proceedings of the IEEE/CVF Conference on Computer Vision and Pattern Recognition}}. \bibinfo{pages}{11453--11464}.
\newblock


\bibitem[Noguchi et~al\mbox{.}(2022)]%
        {enarfgan}
\bibfield{author}{\bibinfo{person}{Atsuhiro Noguchi}, \bibinfo{person}{Xiao Sun}, \bibinfo{person}{Stephen Lin}, {and} \bibinfo{person}{Tatsuya Harada}.} \bibinfo{year}{2022}\natexlab{}.
\newblock \showarticletitle{Unsupervised Learning of Efficient Geometry-Aware Neural Articulated Representations}.
\newblock \bibinfo{journal}{\emph{arXiv preprint arXiv:2204.08839}} (\bibinfo{year}{2022}).
\newblock


\bibitem[Or-El et~al\mbox{.}(2022)]%
        {stylesdf}
\bibfield{author}{\bibinfo{person}{Roy Or-El}, \bibinfo{person}{Xuan Luo}, \bibinfo{person}{Mengyi Shan}, \bibinfo{person}{Eli Shechtman}, \bibinfo{person}{Jeong~Joon Park}, {and} \bibinfo{person}{Ira Kemelmacher-Shlizerman}.} \bibinfo{year}{2022}\natexlab{}.
\newblock \showarticletitle{Stylesdf: High-resolution 3d-consistent image and geometry generation}. In \bibinfo{booktitle}{\emph{Proceedings of the IEEE/CVF Conference on Computer Vision and Pattern Recognition}}. \bibinfo{pages}{13503--13513}.
\newblock


\bibitem[Pan et~al\mbox{.}(2023)]%
        {pan2023drag}
\bibfield{author}{\bibinfo{person}{Xingang Pan}, \bibinfo{person}{Ayush Tewari}, \bibinfo{person}{Thomas Leimk{\"u}hler}, \bibinfo{person}{Lingjie Liu}, \bibinfo{person}{Abhimitra Meka}, {and} \bibinfo{person}{Christian Theobalt}.} \bibinfo{year}{2023}\natexlab{}.
\newblock \showarticletitle{Drag your gan: Interactive point-based manipulation on the generative image manifold}. In \bibinfo{booktitle}{\emph{ACM SIGGRAPH 2023 Conference Proceedings}}. \bibinfo{pages}{1--11}.
\newblock


\bibitem[Qiu et~al\mbox{.}(2023)]%
        {Qiu2023RECMVR3}
\bibfield{author}{\bibinfo{person}{Lingteng Qiu}, \bibinfo{person}{Guanying Chen}, \bibinfo{person}{Jiapeng Zhou}, \bibinfo{person}{Mutian Xu}, \bibinfo{person}{Junle Wang}, {and} \bibinfo{person}{Xiaoguang Han}.} \bibinfo{year}{2023}\natexlab{}.
\newblock \showarticletitle{REC-MV: REconstructing 3D Dynamic Cloth from Monocular Videos}.
\newblock \bibinfo{journal}{\emph{2023 IEEE/CVF Conference on Computer Vision and Pattern Recognition (CVPR)}} (\bibinfo{year}{2023}), \bibinfo{pages}{4637--4646}.
\newblock
\urldef\tempurl%
\url{https://api.semanticscholar.org/CorpusID:258841295}
\showURL{%
\tempurl}


\bibitem[Radford et~al\mbox{.}(2021)]%
        {clip}
\bibfield{author}{\bibinfo{person}{Alec Radford}, \bibinfo{person}{Jong~Wook Kim}, \bibinfo{person}{Chris Hallacy}, \bibinfo{person}{Aditya Ramesh}, \bibinfo{person}{Gabriel Goh}, \bibinfo{person}{Sandhini Agarwal}, \bibinfo{person}{Girish Sastry}, \bibinfo{person}{Amanda Askell}, \bibinfo{person}{Pamela Mishkin}, \bibinfo{person}{Jack Clark}, {et~al\mbox{.}}} \bibinfo{year}{2021}\natexlab{}.
\newblock \showarticletitle{Learning transferable visual models from natural language supervision}. In \bibinfo{booktitle}{\emph{International conference on machine learning}}. PMLR, \bibinfo{pages}{8748--8763}.
\newblock


\bibitem[Rombach et~al\mbox{.}(2022)]%
        {ldm}
\bibfield{author}{\bibinfo{person}{Robin Rombach}, \bibinfo{person}{Andreas Blattmann}, \bibinfo{person}{Dominik Lorenz}, \bibinfo{person}{Patrick Esser}, {and} \bibinfo{person}{Bj{\"o}rn Ommer}.} \bibinfo{year}{2022}\natexlab{}.
\newblock \showarticletitle{High-resolution image synthesis with latent diffusion models}. In \bibinfo{booktitle}{\emph{Proceedings of the IEEE/CVF conference on computer vision and pattern recognition}}. \bibinfo{pages}{10684--10695}.
\newblock


\bibitem[Sarkar et~al\mbox{.}(2021a)]%
        {sarkar2021style}
\bibfield{author}{\bibinfo{person}{Kripasindhu Sarkar}, \bibinfo{person}{Vladislav Golyanik}, \bibinfo{person}{Lingjie Liu}, {and} \bibinfo{person}{Christian Theobalt}.} \bibinfo{year}{2021}\natexlab{a}.
\newblock \showarticletitle{Style and pose control for image synthesis of humans from a single monocular view}.
\newblock \bibinfo{journal}{\emph{arXiv preprint arXiv:2102.11263}} (\bibinfo{year}{2021}).
\newblock


\bibitem[Sarkar et~al\mbox{.}(2021b)]%
        {humangan}
\bibfield{author}{\bibinfo{person}{Kripasindhu Sarkar}, \bibinfo{person}{Lingjie Liu}, \bibinfo{person}{Vladislav Golyanik}, {and} \bibinfo{person}{Christian Theobalt}.} \bibinfo{year}{2021}\natexlab{b}.
\newblock \showarticletitle{HumanGAN: A Generative Model of Humans Images}.
\newblock \bibinfo{journal}{\emph{arXiv preprint arXiv:2103.06902}} (\bibinfo{year}{2021}).
\newblock


\bibitem[Schroff et~al\mbox{.}(2015)]%
        {Schroff2015FaceNetAU}
\bibfield{author}{\bibinfo{person}{Florian Schroff}, \bibinfo{person}{Dmitry Kalenichenko}, {and} \bibinfo{person}{James Philbin}.} \bibinfo{year}{2015}\natexlab{}.
\newblock \showarticletitle{FaceNet: A unified embedding for face recognition and clustering}.
\newblock \bibinfo{journal}{\emph{2015 IEEE Conference on Computer Vision and Pattern Recognition (CVPR)}} (\bibinfo{year}{2015}), \bibinfo{pages}{815--823}.
\newblock
\urldef\tempurl%
\url{https://api.semanticscholar.org/CorpusID:206592766}
\showURL{%
\tempurl}


\bibitem[Shen and Zhou(2021)]%
        {shen2021closed}
\bibfield{author}{\bibinfo{person}{Yujun Shen} {and} \bibinfo{person}{Bolei Zhou}.} \bibinfo{year}{2021}\natexlab{}.
\newblock \showarticletitle{Closed-form factorization of latent semantics in gans}. In \bibinfo{booktitle}{\emph{Proceedings of the IEEE/CVF conference on computer vision and pattern recognition}}. \bibinfo{pages}{1532--1540}.
\newblock


\bibitem[Shue et~al\mbox{.}(2022)]%
        {shue20223d}
\bibfield{author}{\bibinfo{person}{J~Ryan Shue}, \bibinfo{person}{Eric~Ryan Chan}, \bibinfo{person}{Ryan Po}, \bibinfo{person}{Zachary Ankner}, \bibinfo{person}{Jiajun Wu}, {and} \bibinfo{person}{Gordon Wetzstein}.} \bibinfo{year}{2022}\natexlab{}.
\newblock \showarticletitle{3D Neural Field Generation using Triplane Diffusion}.
\newblock \bibinfo{journal}{\emph{arXiv preprint arXiv:2211.16677}} (\bibinfo{year}{2022}).
\newblock


\bibitem[Su et~al\mbox{.}(2022)]%
        {Su2022DrawingInStylesPI}
\bibfield{author}{\bibinfo{person}{Wanchao Su}, \bibinfo{person}{Hui Ye}, \bibinfo{person}{Shu-Yu Chen}, \bibinfo{person}{Lin Gao}, {and} \bibinfo{person}{Hongbo Fu}.} \bibinfo{year}{2022}\natexlab{}.
\newblock \showarticletitle{DrawingInStyles: Portrait Image Generation and Editing with Spatially Conditioned StyleGAN}.
\newblock \bibinfo{journal}{\emph{IEEE Transactions on Visualization and Computer Graphics}}  \bibinfo{volume}{PP} (\bibinfo{year}{2022}), \bibinfo{pages}{1--1}.
\newblock
\urldef\tempurl%
\url{https://api.semanticscholar.org/CorpusID:247292402}
\showURL{%
\tempurl}


\bibitem[Voynov and Babenko(2020)]%
        {voynov2020unsupervised}
\bibfield{author}{\bibinfo{person}{Andrey Voynov} {and} \bibinfo{person}{Artem Babenko}.} \bibinfo{year}{2020}\natexlab{}.
\newblock \showarticletitle{Unsupervised discovery of interpretable directions in the gan latent space}. In \bibinfo{booktitle}{\emph{International conference on machine learning}}. PMLR, \bibinfo{pages}{9786--9796}.
\newblock


\bibitem[Wang et~al\mbox{.}(2022)]%
        {wang2022rodin}
\bibfield{author}{\bibinfo{person}{Tengfei Wang}, \bibinfo{person}{Bo Zhang}, \bibinfo{person}{Ting Zhang}, \bibinfo{person}{Shuyang Gu}, \bibinfo{person}{Jianmin Bao}, \bibinfo{person}{Tadas Baltrusaitis}, \bibinfo{person}{Jingjing Shen}, \bibinfo{person}{Dong Chen}, \bibinfo{person}{Fang Wen}, \bibinfo{person}{Qifeng Chen}, {et~al\mbox{.}}} \bibinfo{year}{2022}\natexlab{}.
\newblock \showarticletitle{Rodin: A Generative Model for Sculpting 3D Digital Avatars Using Diffusion}.
\newblock \bibinfo{journal}{\emph{arXiv preprint arXiv:2212.06135}} (\bibinfo{year}{2022}).
\newblock


\bibitem[Wang et~al\mbox{.}(2004)]%
        {ssim}
\bibfield{author}{\bibinfo{person}{Zhou Wang}, \bibinfo{person}{A. Bovik}, \bibinfo{person}{H.~R. Sheikh}, {and} \bibinfo{person}{E.~P. Simoncelli}.} \bibinfo{year}{2004}\natexlab{}.
\newblock \showarticletitle{Image quality assessment: from error visibility to structural similarity}.
\newblock \bibinfo{journal}{\emph{IEEE Transactions on Image Processing}}  \bibinfo{volume}{13} (\bibinfo{year}{2004}), \bibinfo{pages}{600--612}.
\newblock


\bibitem[Wolff et~al\mbox{.}(2021)]%
        {Wolff2021DesigningPG}
\bibfield{author}{\bibinfo{person}{Katja Wolff}, \bibinfo{person}{Philipp Herholz}, \bibinfo{person}{Verena Ziegler}, \bibinfo{person}{Frauke Link}, \bibinfo{person}{Nico Br{\"u}gel}, {and} \bibinfo{person}{Olga Sorkine-Hornung}.} \bibinfo{year}{2021}\natexlab{}.
\newblock \showarticletitle{Designing Personalized Garments with Body Movement}.
\newblock \bibinfo{journal}{\emph{Computer Graphics Forum}}  \bibinfo{volume}{42} (\bibinfo{year}{2021}).
\newblock
\urldef\tempurl%
\url{https://api.semanticscholar.org/CorpusID:252383414}
\showURL{%
\tempurl}


\bibitem[Wu et~al\mbox{.}(2022)]%
        {Wu2022DeepPortraitDrawingGH}
\bibfield{author}{\bibinfo{person}{X. Wu}, \bibinfo{person}{Chen Wang}, \bibinfo{person}{Hongbo Fu}, \bibinfo{person}{Ariel Shamir}, \bibinfo{person}{Song-Hai Zhang}, {and} \bibinfo{person}{Shimin Hu}.} \bibinfo{year}{2022}\natexlab{}.
\newblock \showarticletitle{DeepPortraitDrawing: Generating Human Body Images from Freehand Sketches}.
\newblock \bibinfo{journal}{\emph{Comput. Graph.}}  \bibinfo{volume}{116} (\bibinfo{year}{2022}), \bibinfo{pages}{73--81}.
\newblock
\urldef\tempurl%
\url{https://api.semanticscholar.org/CorpusID:248512601}
\showURL{%
\tempurl}


\bibitem[Xiong et~al\mbox{.}(2023)]%
        {Xiong2023Get3DHumanLS}
\bibfield{author}{\bibinfo{person}{Zhangyang Xiong}, \bibinfo{person}{Di Kang}, \bibinfo{person}{Derong Jin}, \bibinfo{person}{Weikai Chen}, \bibinfo{person}{Linchao Bao}, {and} \bibinfo{person}{Xiaoguang Han}.} \bibinfo{year}{2023}\natexlab{}.
\newblock \showarticletitle{Get3DHuman: Lifting StyleGAN-Human into a 3D Generative Model using Pixel-aligned Reconstruction Priors}.
\newblock \bibinfo{journal}{\emph{2023 IEEE/CVF International Conference on Computer Vision (ICCV)}} (\bibinfo{year}{2023}), \bibinfo{pages}{9253--9263}.
\newblock
\urldef\tempurl%
\url{https://api.semanticscholar.org/CorpusID:256503649}
\showURL{%
\tempurl}


\bibitem[Zablotskaia et~al\mbox{.}(2019)]%
        {ubcfashion}
\bibfield{author}{\bibinfo{person}{Polina Zablotskaia}, \bibinfo{person}{Aliaksandr Siarohin}, \bibinfo{person}{Bo Zhao}, {and} \bibinfo{person}{Leonid Sigal}.} \bibinfo{year}{2019}\natexlab{}.
\newblock \showarticletitle{Dwnet: Dense warp-based network for pose-guided human video generation}.
\newblock \bibinfo{journal}{\emph{arXiv preprint arXiv:1910.09139}} (\bibinfo{year}{2019}).
\newblock


\bibitem[Zeng et~al\mbox{.}(2020)]%
        {Zeng20203DHM}
\bibfield{author}{\bibinfo{person}{Wang Zeng}, \bibinfo{person}{Wanli Ouyang}, \bibinfo{person}{Ping Luo}, \bibinfo{person}{Wentao Liu}, {and} \bibinfo{person}{Xiaogang Wang}.} \bibinfo{year}{2020}\natexlab{}.
\newblock \showarticletitle{3D Human Mesh Regression With Dense Correspondence}.
\newblock \bibinfo{journal}{\emph{2020 IEEE/CVF Conference on Computer Vision and Pattern Recognition (CVPR)}} (\bibinfo{year}{2020}), \bibinfo{pages}{7052--7061}.
\newblock
\urldef\tempurl%
\url{https://api.semanticscholar.org/CorpusID:219558352}
\showURL{%
\tempurl}


\bibitem[Zeng et~al\mbox{.}(2022)]%
        {zeng2022lion}
\bibfield{author}{\bibinfo{person}{Xiaohui Zeng}, \bibinfo{person}{Arash Vahdat}, \bibinfo{person}{Francis Williams}, \bibinfo{person}{Zan Gojcic}, \bibinfo{person}{Or Litany}, \bibinfo{person}{Sanja Fidler}, {and} \bibinfo{person}{Karsten Kreis}.} \bibinfo{year}{2022}\natexlab{}.
\newblock \showarticletitle{LION: Latent Point Diffusion Models for 3D Shape Generation}.
\newblock \bibinfo{journal}{\emph{arXiv preprint arXiv:2210.06978}} (\bibinfo{year}{2022}).
\newblock


\bibitem[Zhang et~al\mbox{.}(2023)]%
        {controlnet}
\bibfield{author}{\bibinfo{person}{Lvmin Zhang}, \bibinfo{person}{Anyi Rao}, {and} \bibinfo{person}{Maneesh Agrawala}.} \bibinfo{year}{2023}\natexlab{}.
\newblock \bibinfo{title}{Adding Conditional Control to Text-to-Image Diffusion Models}.
\newblock
\newblock


\bibitem[Zhang et~al\mbox{.}(2018)]%
        {lpip}
\bibfield{author}{\bibinfo{person}{Richard Zhang}, \bibinfo{person}{Phillip Isola}, \bibinfo{person}{Alexei~A. Efros}, \bibinfo{person}{E. Shechtman}, {and} \bibinfo{person}{O. Wang}.} \bibinfo{year}{2018}\natexlab{}.
\newblock \showarticletitle{The Unreasonable Effectiveness of Deep Features as a Perceptual Metric}.
\newblock \bibinfo{journal}{\emph{CVPR}} (\bibinfo{year}{2018}), \bibinfo{pages}{586--595}.
\newblock


\bibitem[Zhou et~al\mbox{.}(2021)]%
        {zhou20213d}
\bibfield{author}{\bibinfo{person}{Linqi Zhou}, \bibinfo{person}{Yilun Du}, {and} \bibinfo{person}{Jiajun Wu}.} \bibinfo{year}{2021}\natexlab{}.
\newblock \showarticletitle{3d shape generation and completion through point-voxel diffusion}. In \bibinfo{booktitle}{\emph{Proceedings of the IEEE/CVF International Conference on Computer Vision}}. \bibinfo{pages}{5826--5835}.
\newblock


\bibitem[Zhu et~al\mbox{.}(2020)]%
        {Zhu2020DeepFA}
\bibfield{author}{\bibinfo{person}{Heming Zhu}, \bibinfo{person}{Yu Cao}, \bibinfo{person}{Hang Jin}, \bibinfo{person}{Weikai Chen}, \bibinfo{person}{Dong Du}, \bibinfo{person}{Zhangye Wang}, \bibinfo{person}{Shuguang Cui}, {and} \bibinfo{person}{Xiaoguang Han}.} \bibinfo{year}{2020}\natexlab{}.
\newblock \showarticletitle{Deep Fashion3D: A Dataset and Benchmark for 3D Garment Reconstruction from Single Images}.
\newblock \bibinfo{journal}{\emph{ArXiv}}  \bibinfo{volume}{abs/2003.12753} (\bibinfo{year}{2020}).
\newblock
\urldef\tempurl%
\url{https://api.semanticscholar.org/CorpusID:214714400}
\showURL{%
\tempurl}


\bibitem[Zhu et~al\mbox{.}(2022)]%
        {Zhu2022RegisteringET}
\bibfield{author}{\bibinfo{person}{Heming Zhu}, \bibinfo{person}{Lingteng Qiu}, \bibinfo{person}{Yuda Qiu}, {and} \bibinfo{person}{Xiaoguang Han}.} \bibinfo{year}{2022}\natexlab{}.
\newblock \showarticletitle{Registering Explicit to Implicit: Towards High-Fidelity Garment mesh Reconstruction from Single Images}.
\newblock \bibinfo{journal}{\emph{2022 IEEE/CVF Conference on Computer Vision and Pattern Recognition (CVPR)}} (\bibinfo{year}{2022}), \bibinfo{pages}{3835--3844}.
\newblock
\urldef\tempurl%
\url{https://api.semanticscholar.org/CorpusID:247778386}
\showURL{%
\tempurl}


\bibitem[Zhu et~al\mbox{.}(2016)]%
        {zhu2016generative}
\bibfield{author}{\bibinfo{person}{Jun-Yan Zhu}, \bibinfo{person}{Philipp Kr{\"a}henb{\"u}hl}, \bibinfo{person}{Eli Shechtman}, {and} \bibinfo{person}{Alexei~A Efros}.} \bibinfo{year}{2016}\natexlab{}.
\newblock \showarticletitle{Generative visual manipulation on the natural image manifold}. In \bibinfo{booktitle}{\emph{Computer Vision--ECCV 2016: 14th European Conference, Amsterdam, The Netherlands, October 11-14, 2016, Proceedings, Part V 14}}. Springer, \bibinfo{pages}{597--613}.
\newblock


\end{thebibliography}

\end{document}